\documentclass[lettersize,journal]{IEEEtran}
\usepackage{amsmath,amsfonts}
\usepackage{algorithmic}
\usepackage{algorithm}
\usepackage{array}
\usepackage{textcomp}
\usepackage{stfloats}
\usepackage{url}
\usepackage{verbatim}
\usepackage{graphicx}
\usepackage{cite}
\usepackage{booktabs}
\usepackage{xspace}
\usepackage{microtype}

\usepackage{graphicx}
\usepackage{subcaption}
\usepackage[pagebackref,breaklinks,colorlinks]{hyperref}

\usepackage{xcolor}
\definecolor{Graylight}{gray}{0.9}
\definecolor{Gray}{gray}{1.0}
\usepackage{colortbl}
\usepackage{multirow}
\usepackage{makecell}
\usepackage{bm}
\usepackage{pifont}
\definecolor{mygreen}{RGB}{9,136,66}

\usepackage{url}
\usepackage{graphicx}
\usepackage{xspace}
\usepackage{colortbl}
\usepackage{booktabs}
\usepackage{multirow,marvosym}

\usepackage{xcolor}

\definecolor{ForestGreen}{rgb}{0, 0.69, 0.31}
\definecolor{NavyBlue}{rgb}{0, 0.44, 0.75}
\newcommand{\hgreen}[1]{\textcolor{ForestGreen}{\textbf{#1}}} 
\newcommand{\hblue}[1]{\textcolor{NavyBlue}{\textbf{#1}}} 

\usepackage{pifont}

\newcommand{\xmark}{\ding{55}}
\newcommand{\methodname}{Visual-RFT++\xspace}
\newcommand{\codename}{MAT-Coding\xspace}
\newcommand{\searchname}{MAT-Search\xspace}
\newcommand{\benchname}{MAT\xspace}
\definecolor{Cerulean}{rgb}{0.0, 0.48, 0.65}

\definecolor{Cerulean}{rgb}{0.0, 0.48, 0.65}

\usepackage{amsmath}

\begin{document}

\title{Visual-RFT++: Extending Visual Reinforcement Fine-Tuning to Multimodal Tool Use}

\author{Ziyu Liu, Zeyi Sun, Xiaoyi Dong, Yuhang Cao, Haodong Duan, \\
Dahua Lin, Jiaqi Wang, Yuhang Zang\textsuperscript{\Letter}
\thanks{Ziyu Liu and Zeyi Sun are with Shanghai Jiao Tong University.}
\thanks{Yuhang Zang, Xiaoyi Dong, Yuhang Cao, Haodong Duan, Jiaqi Wang are with Shanghai AI Laboratory.}
\thanks{Dahua Lin is with The Chinese University of Hong Kong.}
\thanks{Corresponding author: Yuhang Zang (zangyuhang@pjlab.org.cn).}
}

\markboth{Journal of \LaTeX\ Class Files,~Vol.~14, No.~8, August~2021}%
{Shell \MakeLowercase{\textit{et al.}}: A Sample Article Using IEEEtran.cls for IEEE Journals}


\maketitle

\begin{abstract}
Reinforcement Fine-Tuning (RFT) has recently become a central paradigm for advancing Large Reasoning Models (LRMs), as demonstrated by systems such as OpenAI’s o3 and the R1 series. Despite significant progress in LLMs, how to leverage RFT to unify visual perception, step-by-step reasoning, and agentic tool use into one coherent training paradigm remains an open challenge.
In this work, we present Visual Reinforcement Fine-Tuning++ (\methodname), a unified framework that extends RFT to both visual perception tasks and multimodal agentic workflows, enabling LVLMs to reason with images, act upon images, and evaluate their own outputs using verifiable reward.
With \methodname, open-source LVLMs not only show significant gains on diverse visual perception tasks, but also acquire agentic abilities such as web searching for real-time information and code-based image manipulation (e.g., cropping, rotation, and analysis).
Across visual perception tasks, including fine-grained classification, few-shot detection, reasoning grounding, and open-vocabulary detection, \methodname delivers strong gains over SFT, including a +24.3 percentage points improvement in one-shot fine-grained classification and +21.0 percentage points on COCO's 4-shot setting and +15.4 percentage points on LVIS with around several hundred samples.
To evaluate agentic behaviors, we introduce the Multimodal Agentic Tool Bench (MAT) with MAT-Search and MAT-Coding. \methodname surpasses its baseline by +18.6 F1 / +13.0 EM percentage points on MAT-Coding and +10.3 F1 / +8.7 EM percentage points on MAT-Search.
It also shows gains of 29.3 F1 and 25.9 EM percentage points on multi-hop QA benchmarks such as 2Wiki and HotpotQA, demonstrating strong cross-modal generalization.
\methodname introduces a data-efficient and reward-driven framework that strengthens both visual reasoning and agentic adaptability.
Our code is available at \href{https://github.com/Liuziyu77/Visual-RFT}{https://github.com/Liuziyu77/Visual-RFT}.
\end{abstract}


\section{Introduction}
\begin{figure}[t]
    \begin{center}
    \vspace{-12pt}
    \includegraphics[width=1.0\linewidth]{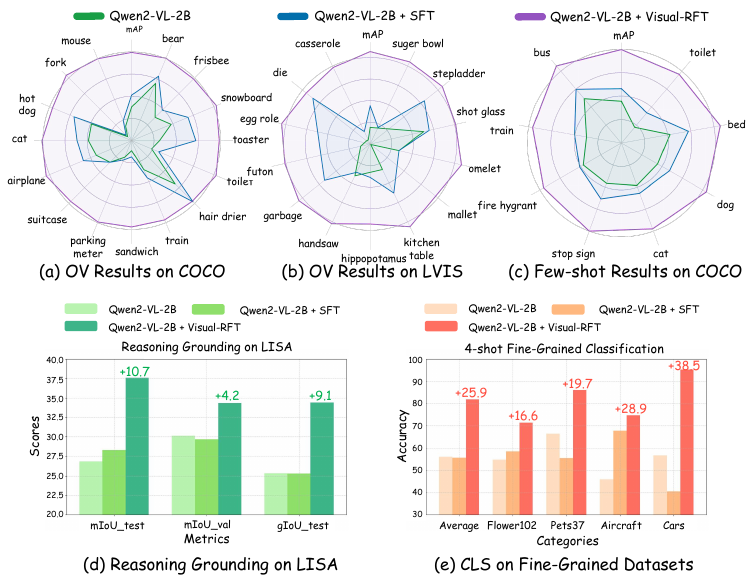}
    \end{center}
    \vspace{-12pt}
    \caption{\small Our \textbf{Visual} \textbf{R}einforcement \textbf{F}ine-\textbf{T}uning++ (\methodname) performs better than previous Supervised Fine-Tuning (SFT) on a variety of tasks, such as Open Vocabulary(OV)/Few-shot Detection, Reasoning Grounding, and Fine-grained Classification.}
    \label{fig:teaser}
    \vspace{-12pt}
\end{figure}

\begin{figure*}[t]
    \begin{center}
    \vspace{-24pt}
    \includegraphics[width=1.\linewidth]{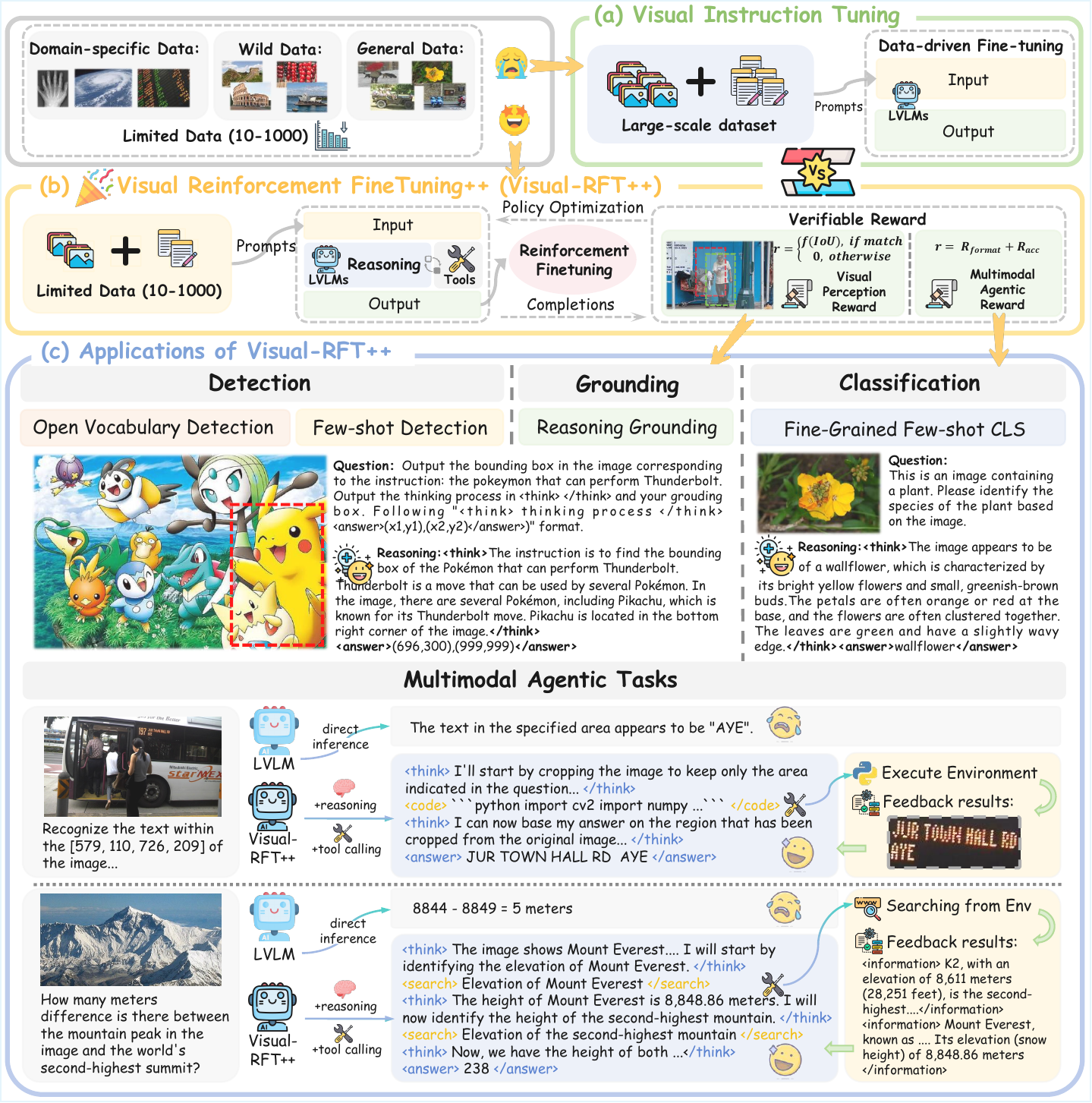}
    \end{center}
    \vspace{-14pt}
    \caption{\small \textbf{Overview of \methodname.}  Compared to the \textbf{(a)} Visual Instruction Tuning that is data-hungry, \textbf{(b)} our Visual Reinforcement Fine-Tuning++ (\methodname) is more data efficient with limited data.
    \textbf{(c)} We successfully empower Large Vision-Language Models (LVLMs) with RFT on a series of multimodal tasks, and present examples of the model's reasoning process at the bottom.}
    \label{fig:teaser_framework}
    \vspace{-12pt}
\end{figure*}

Large Reasoning Models (LRMs) such as OpenAI o1~\cite{OpenAI_O1} and o3~\cite{OpenAI-o3} represent the frontier of AI models that spend more computation “thinking’’ before answering, exhibiting strong multi-step reasoning and tool-augmented decision-making abilities across both textual and visual modalities. A key factor behind these advancements is Reinforcement Fine-Tuning (RFT)\footnote{\href{https://openai.com/form/rft-research-program}{https://openai.com/form/rft-research-program}}, which enables models to adapt to new domains with only dozens to thousands of examples through reward-driven learning.
Although the implementation details of o1 and o3 remain proprietary, open-source efforts such as DeepSeek R1~\cite{DeepSeek-R1} suggest that verifiable rewards~\cite{lambert2024ttulu,team2025kimi}, which rely on rule-based correctness checks rather than learned preference models~\cite{ouyang2022training,liu2024skywork,zang2025internlm}, play an essential role in reproducing RFT behavior.

A major advantage of RFT over traditional Supervised Fine-Tuning (SFT) lies in its data efficiency.
 SFT (see Fig.~\ref{fig:teaser_framework} (a)) requires curated “ground-truth’’ annotations and large-scale datasets, while RFT evaluates the model’s own responses and updates its policy depending on correctness, enabling effective learning through trial and error~\cite{OpenAI_O1,OpenAI_O3}.
A commonly held belief is that RFT applies only to tasks such as math and code generation.
Math and coding exhibit clear and objective final answers or test cases, making their rewards relatively straightforward to verify.
However, recent developments in multimodal LRMs highlight a growing need for RFT not only for visual perception but also for agentic decision-making with tools, such as web browsing~\cite{li2025webthinker}, code execution~\cite{li2025torl}, and image manipulation~\cite{OpenAI-o3}.

Motivated by this emerging trend, we revisit the role of RFT in multimodal settings and demonstrate that it can be extended far beyond math and code, enabling models to learn both visual perception and visual agentic behaviors through verifiable reward. To this end, we propose \textbf{Visual} \textbf{R}einforcement \textbf{F}ine-\textbf{T}uning++ (\methodname), which empowers Large Vision-Language Models (LVLMs) with both visual perception capabilities (e.g., fine-grained classification, grounding, few-shot detection, open-vocabulary detection) and agentic tool-use abilities (e.g., web search, code-based image understanding), as shown in Fig.~\ref{fig:teaser}.

To extend RFT on visual tasks, we present the implementation details of \methodname in Fig.~\ref{fig:teaser_framework} (b).
For each input, \methodname first samples multiple trajectories from an LVLM, each containing structured reasoning and a final answer. We then apply task-specific rule-based verifiable rewards to guide policy learning with algorithms such as GRPO~\cite{grpo}.
For instance, we propose the Intersection over Union (IoU) reward for the object detection task.
For agentic tasks, we further define rewards that verify tool-use behaviors.
Rather than memorizing annotated answers as in SFT, \methodname encourages LVLMs to explore solutions, reason step-by-step, invoke tools when needed, and optimize for outcomes aligned with our verifiable reward functions.
The goal is to discover optimal solutions, rather than merely mimicking pre-defined answers.
Our approach shifts the training paradigm from data scaling in SFT to the strategic design of varied reward functions tailored to specific multimodal tasks.
As shown in Fig.~\ref{fig:teaser_framework} (c), the synergistic combination of verifiable rewards and visual perception abilities (e.g., detection, grounding, classification, agentic tasks) allows our model to achieve rapid and data-efficient mastery of new concepts, facilitated by a detailed reasoning process.

We validate the effectiveness of \methodname across a broad spectrum of visual perception and multimodal agentic tasks.
For visual perception, \methodname achieves substantial improvements over SFT on fine-grained classification, few-shot detection, reasoning grounding, and open-vocabulary detection. In the one-shot fine-grained classification setting with only a few hundred samples, \methodname improves accuracy by 24.3 percentage points while SFT exhibits a 4.3-percentage-point drop. In few-shot object detection, \methodname also demonstrates strong data efficiency, achieving mAP gains of 21.0 percentage points on COCO four-shot and 15.4 percentage points on LVIS. On the LISA dataset~\cite{lai2024lisa}, which requires reasoning-grounded localization, \methodname surpasses specialized models such as GroundedSAM~\cite{liu2024grounding}. Moreover, \methodname generalizes effectively to new and rare categories in open-vocabulary detection, with the 2B model improving nAP from 9.6 to 27.9 on COCO~\cite{coco} new classes and from 2.7 to 20.7 on LVIS~\cite{gupta2019lvis} rare categories.

Beyond perception tasks, \methodname also equips LVLMs with strong multimodal agentic abilities. To facilitate training and evaluation, we introduce the \textbf{M}ultimodal \textbf{A}gentic \textbf{T}ool Bench (\benchname), which includes \searchname and \codename. \benchname is designed to measure agentic reasoning accuracy, tool execution correctness, and multimodal generalization. On \codename and \searchname, \methodname achieves competitive performance. Notably, the Qwen2.5-VL-3B model achieves higher F1 than direct-inference GPT-4o on MAT-Coding task, and the Qwen2.5-VL-7B model improves upon the baseline by 18.56 F1 and 13.00 EM percentage points. On the \searchname task, the 7B model further achieves gains of 10.28 F1 and 8.66 EM percentage points. In addition, \methodname generalizes strongly to out-of-domain multi-hop QA datasets, exhibiting consistent improvements across challenging benchmarks.

Together, these results demonstrate that \methodname provides a unified and data-efficient approach that enhances visual perception and multimodal agentic behaviors. Here, ``unified'' refers to a common GRPO-based reinforcement fine-tuning methodology with task-specific verifiable rewards.






A preliminary version of this work (Visual-RFT~\cite{liu2025visual}) was accepted at ICCV 2025.
In this extended version, we substantially expand the original framework into \methodname, upgrading the scope, methodology, and experimental coverage in several key dimensions:
\begin{itemize}
\item \methodname extends the use of RFT from perception-only settings to multimodal agentic reasoning, enabling LVLMs to perform tool-augmented tasks such as task decomposition, web-based retrieval and code-driven image manipulation.
\item We introduce a more comprehensive and modular reward design, covering both perception correctness signals and verification mechanisms tailored for agentic tasks, offering a unified verifiable-reward framework for diverse multimodal tasks.
\item We develop the Multimodal Agentic Tool Bench (MAT), a new benchmark suite consisting of MAT-Search and MAT-Coding, which systematically evaluates tool-use reasoning and multimodal agentic capabilities not studied in the original Visual-RFT.
\item We provide substantially expanded experiments, including evaluations on visual perception, multimodal agentic tasks, and out-of-domain multi-hop QA, showing that \methodname achieves significant improvements over supervised fine-tuning and baseline LVLMs.
\item In addition to methodological extensions, we conduct extensive ablation studies to examine how different reward designs affect model performance. We also analyze the advantages of \methodname over other tool-based approaches, highlighting its superior reliability and generalization across diverse multimodal tasks.
\item Moreover, we provide deeper analysis of our reward and pipeline design mechanisms, as well as the data construction process, particularly for multimodal agentic tasks. This analysis offers insights into why RFT enables stronger visual understanding, more reliable agentic reasoning, and how effective RFT algorithms can be designed.
\end{itemize}
\section{Related Works}

\noindent \textbf{Large Vision Language Models} (LVLMs) like GPT-4o~\cite{2024gpt4o} achieve excellent visual understanding ability by integrating both visual and textual data. This integration enhances the models' ability to understand complex multimodal inputs and enables more advanced AI systems~\cite{wang2024qwen2vl,li2024llavaov, liang2024survey, zhang2024vision, LVLM-EHub,10363674, 10356651, 11136157, 10817587, xcomposer2.5,liu2024deepseekv3} capable of processing and responding to both images and text. Generally, the training of LVLMs involves two steps: (a) pre-training and (b) post-training, which contains supervised fine-tuning and reinforcement learning. Post-training is crucial in improving the model's response quality, instruction following ability, and reasoning abilities. While significant research has focused on using reinforcement learning to enhance LLMs during post-training~\cite{lm-human-preferences, Learning-to-summarize, Training-language-models, RL4LMs, zang2024contextual, Grounding-large-language-models, llavarlhf, ILQL, LMRL-Gym, ArCHer, ReAct}, the progress for LVLMs has been slower. In this paper, we propose \methodname.
Our method uses GRPO-based reinforcement optimization and verifiable rewards to enhance not only the model's visual perception and reasoning abilities, but also its multimodal agentic capabilities.

\noindent \textbf{Reinforcement Learning} Recently, with the emergence of reasoning models like OpenAI's o1~\cite{OpenAI_O1}, the research focus in Large Language Models (LLMs) has increasingly shifted towards enhancing the models' reasoning capabilities through reinforcement learning (RL) techniques. Studies have explored improving LLMs' performance in reasoning tasks such as solving mathematical problems~\cite{grpo,yang2024qwen2math,ying2024internlmmath,cai2024internlm2,luong2024reft} and coding~\cite{hui2024qwen2coder,jiao2024preferencecode,zhang2024o1,zhang2024codedpo}. A notable breakthrough in this area is Deepseek-R1-Zero~\cite{DeepSeek-R1}, which introduced a new approach to achieving robust reasoning capabilities using only RL, eliminating the supervised fine-tuning (SFT) stage.
Recent studies have extended reinforcement learning with verifiable rewards to LVLMs. R1-V~\cite{chen2025r1v}, VLM-R1~\cite{shen2025vlmr1}, Vision-R1~\cite{Huang2025VisionR1IR}, Seg-Zero~\cite{Liu2025SegZeroRG}, MM-Eureka~\cite{Meng2025MMEurekaET}, MMR1~\cite{Leng2025MMR1EM}, LMM-R1~\cite{Peng2025LMMR1E3}, OpenVLThinker~\cite{Hu2026OpenVLThinkerV2AG}, Curr-ReFT~\cite{Deng2025BoostingTG}, and Reason-RFT~\cite{Tan2025ReasonRFTRF} collectively investigate multimodal VQA, mathematical reasoning, grounding, and segmentation. Although these methods share GRPO-style optimization or verifiable rewards, their task definitions, output spaces, supervision scales, and evaluation protocols differ substantially. Building on Visual-RFT~\cite{liu2025visualrft}, Visual-RFT++ complements these efforts by studying fine-grained classification, few-shot detection, reasoning grounding, and open-vocabulary detection under a unified reward formulation, and further extending the framework to multimodal agentic settings.


\noindent \textbf{Multimodal Agentic Reinforcement Learning and Evaluation.}
Recent agentic-RL methods train LVLMs to interact with external tools during visual reasoning. OpenThinkIMG~\cite{Su2025OpenThinkIMGLT}, DeepEyes~\cite{Zheng2025DeepEyesI}, Pixel-Reasoner~\cite{Su2025PixelRI}, and VisTA~\cite{Huang2025VisualToolAgentA} primarily study visual operations such as cropping, zooming, and grounding, while GUI-R1~\cite{Luo2025GUIR1A} focuses on actions in graphical user interfaces. ToolRL~\cite{qian2025toolrl} and Nemotron-Research-Tool-N1~\cite{zhang2025nemotron} instead emphasize text-based function calling. Related benchmarks also target different application scenarios: GAIA~\cite{Mialon2023GAIAAB} evaluates general-assistant tasks involving browsing and tool use, AgentBench~\cite{Liu2023AgentBenchEL} covers heterogeneous interactive environments, and VisualWebArena~\cite{Koh2024VisualWebArenaEM} focuses on website interaction. MAT instead targets two multimodal tool-use settings: MAT-Search requires iterative web querying with evidence integration, while MAT-Coding requires image analysis followed by code-based image processing for answering. MAT therefore differs from existing methods and benchmarks in terms of task definitions, application scenarios, and tool spaces. Visual-RFT++ further connects these agentic settings with perception-oriented visual reinforcement fine-tuning.

\section{Methods}

\subsection{Preliminary}

\begin{figure*}[t]
    \begin{center}
    \vspace{-14pt}
    \includegraphics[width=1.\linewidth]{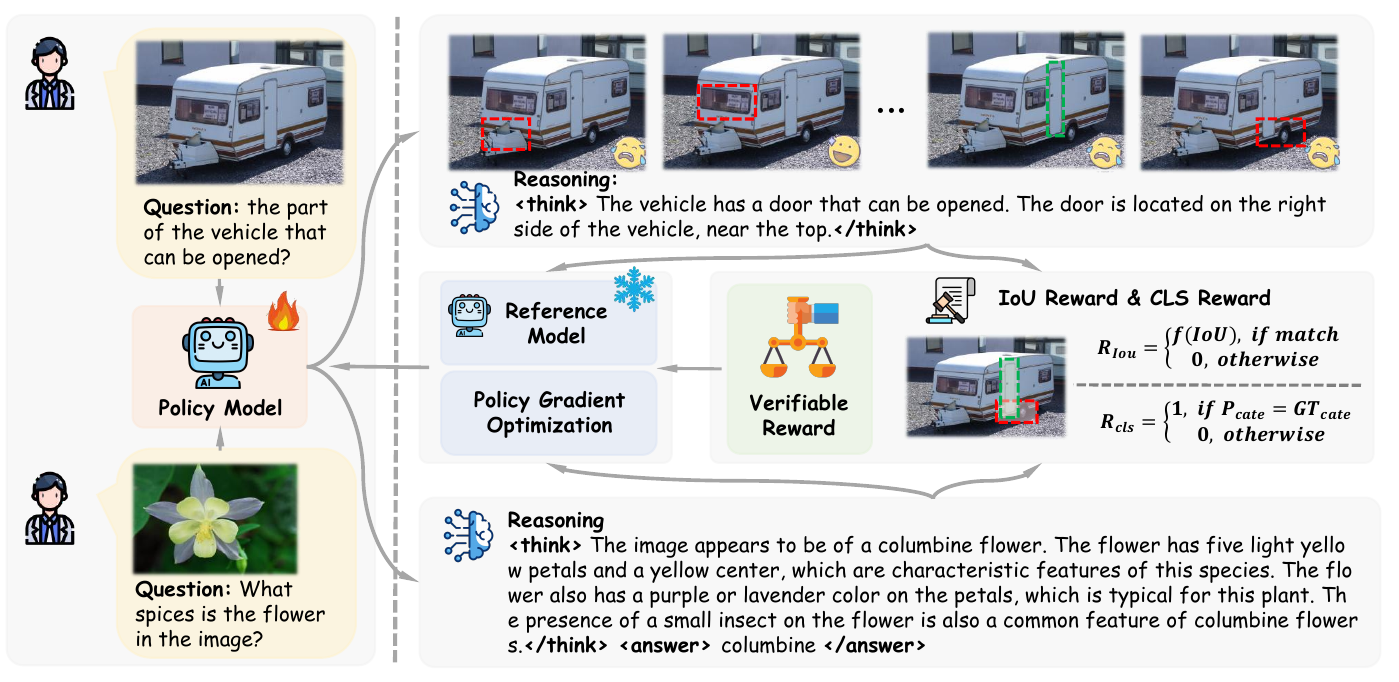}
    \end{center}
    \vspace{-14pt}
    \caption{\small \textbf{Framework of \methodname for Visual Perceptions.} Given the question and visual image inputs, the policy model generates multiple responses containing reasoning steps. Then the verifiable reward such as IoU reward and CLS reward is used with the policy gradient optimization algorithm to update the policy model.}
    \label{fig:framework}
    \vspace{-12pt}
\end{figure*}

\noindent \textbf{Reinforcement Learning with Verifiable Rewards.}
Reinforcement Learning with Verifiable Rewards (RLVR) \cite{lambert2024ttulu,DeepSeek-R1,team2025kimi} is a novel training approach designed to enhance language models in tasks with objectively verifiable outcomes, such as math and coding. Unlike previous Reinforcement Learning from Human Feedback (RLHF) \cite{ouyang2022training,liu2024skywork,zang2025internlm}, which relies on a trained reward model, RLVR uses a direct verification function to assess correctness.
This approach simplifies the reward mechanism while maintaining strong alignment with the task's inherent correctness criteria.

Given the input question $q$, the policy model $\pi_{\theta}$ generates responses $o$ and receives the verifiable reward. More specifically, RLVR optimizes the following objective:
\begin{align}
    & \max_{\pi_{\theta}} \mathbb{E}_{o \sim \pi_{\theta}(q)} \left[ R_{\text{RLVR}}(q, o) \right] \\
    & = \left[ R(q, o) - \beta \text{KL}[\pi_{\theta}(o|q) \parallel \pi_{\text{ref}}(o|q)] \right],
\end{align}
 where $\pi_{\text{ref}}$ is the reference model before optimization, $R$ is the verifiable reward function, and $\beta$ is the hyper-parameter controlling the KL-divergence. The verifiable reward function $R$ takes the question and output pair $(q,o)$ as inputs, and checks if the ground-truth answer matches the prediction $o$:
\begin{equation}
    R(q, o) = 
    \begin{cases} 
    1 ,& \text{if o = ground truth}, \\
    0 ,& \text{otherwise}.
    \end{cases} 
\end{equation}

\noindent \textbf{DeepSeek R1-Zero and GRPO.}
The DeepSeek R1-Zero algorithm eliminates dependence on supervised fine-tuning (SFT) by employing reinforcement learning for training, specifically through its Group Relative Policy Optimization (GRPO) framework. Unlike reinforcement learning algorithms such as PPO \cite{PPO} that require a critic model to evaluate policy performance, GRPO compares groups of candidate responses directly, eliminating the need for an additional critic model. For a given question $q$, GRPO first generates $G$ distinct responses $\{o_1, o_2, ..., o_G\}$ from the current policy $\pi_{\theta_{\text{old}}}$.
Then GRPO takes actions based on these responses and denotes the obtained rewards as $\{r_1, r_2, ..., r_G\}$. By computing their mean and standard deviation for normalization, GRPO determines the relative quality of these responses:
\vspace{-4pt}
\begin{align}
A_i &= \frac{r_i - \text{mean}(\{r_1, \dots, r_G\})}{\text{std}(\{r_1, \dots, r_G\})},
\end{align}
where $A_i$ represents the relative quality of the $i$-th answer. 
GRPO encourages the model to favor better answers with a high reward value within the group.

\subsection{\methodname for Visual Perception}
The framework of \methodname for visual perception is shown in Fig.~\ref{fig:framework}. The multimodal input data from the user consists of images and questions. The policy model $\pi_{\theta}$ outputs a reasoning process and generates a group of responses based on the input. Each response is passed through a verifiable reward function to compute the reward. After group computation of the rewards for each output, the quality of each response is evaluated and used to update the policy model. To ensure the stability of the policy model training, \methodname uses KL divergence to limit the difference between the policy model and the reference model.
We will further discuss how to design the verifiable reward for visual perception tasks in Sec. \ref{sec:reward}, and the data preparation steps in Sec. \ref{sec:data_prepare}.

\subsubsection{Verifiable Reward in Visual Perception}\label{sec:reward}
The reward model is a key component in reinforcement learning (RL) that aligns models with preference alignment algorithms, which can be as straightforward as a verification function that checks for exact matches between predictions and ground-truth answers.
The RL training process in the recent DeepSeek-R1~\cite{DeepSeek-R1} model achieves a significant improvement in the model's reasoning ability through the verifiable reward design.
To transfer this strategy to the visual domain, we design different rule-based verifiable reward functions for various visual perception tasks.

\noindent \textbf{IoU Reward in Detection Tasks.}
For the detection task, the model's output consists of bounding boxes (bboxes) and corresponding confidences.
The reward function for the detection task should adequately consider the Intersection-over-Union (IoU) metric, which is used to compute the mean Average Precision (mAP) in evaluation.
Therefore, we design an IoU and confidence-based reward function $R_d$. First, for the model's output box and confidence, we sort these boxes based on their confidence, denoted as $\{b_1, b_2, ..., b_n\}$. We then match each $b_i$ with the ground truth bbox,$\{b^g_1, b^g_2, ..., b^g_m\}$, and calculate the IoU, while setting an IoU threshold $\tau$. Bounding boxes with an IoU below this threshold $\tau$ are considered invalid, and unmatched bboxes have an IoU of 0. After matching, we obtain the IoU and confidence for each box from the initial set, denoted as $\{iou_1: c_1, iou_2: c_2, ..., iou_n: c_n\}$.

We then use these IoU results and confidence to construct our reward $R_d$. Our reward $R_d$ consists of three parts, including the IoU reward, Confidence reward, and Format reward:
\vspace{-6pt}
\begin{align}
R_{d} = R_{\text{IoU}} + R_{\text{conf}} + R_{\text{format}}.
\end{align}
The IoU reward $R_{\text{IoU}}$ is the average IoU of all the bounding boxes in the model's output,
\vspace{-4pt}
\begin{align}
R_{\text{IoU}} = \frac{iou_1+iou_2+...+iou_n}{n}.
\end{align}
The confidence reward $R_{\text{conf}}$ is related to IoU. For each bounding box, if the $iou_i$ is non-zero, indicating a successful match, the confidence reward for this single box $r_c$ as the predicted confidence is computed as:
\vspace{-4pt}
\begin{align}
r_{ci} = \left\{
\begin{array}{ll}
c_i & , \text{if} \quad iou_i \neq 0,\\
1-c_i & , \text{if} \quad iou_i = 0. \\
\end{array}
\right.
\end{align}
This means that for successfully matched boxes, the higher the confidence, the better. If the $iou_i$ is zero, indicating a failed match, the lower the confidence reward $r_c$ for this box, the better. The overall confidence reward $R_{\text{conf}}$ for the model's output is the average of the confidence rewards of all the bounding boxes in that output,
\begin{align}
R_{\text{conf}} = \frac{ \sum_{i=1}^{n} r_{ci}}{n}.
\end{align}
The format reward $R_{\text{format}}$ is used to encourage the model prediction to adhere to the required HTML tag format of $<$think$>$ and $<$answer$>$ (will be detailed in Sec. \ref{sec:data_prepare}).

\noindent \textbf{CLS Reward in Classification Tasks.}
In classification tasks, the reward function we use consists of two parts: accuracy reward $R_{\text{acc}}$ and format reward $R_{\text{format}}$.
The accuracy reward is determined by comparing the model's output class with the ground truth class, yielding a value of 1 for correct classification and 0 for incorrect classification:
\vspace{-2pt}
\begin{align}
R_{\text{cls}} = R_{\text{acc}} + R_{\text{format}}.
\end{align}

\begin{table}[t]
\caption{
\small
\textbf{Prompts used to construct the dataset.} We have listed the detection prompt and classification prompt separately.}
\vspace{-2mm}
\label{tab:prompt}
\begin{center}
\setlength{\tabcolsep}{4pt}
\scalebox{.95}
{
\begin{tabular}{p{9cm}}
\toprule
\textbf{Detection Prompt:} Detect all objects belonging to the category '$\{$category$\}$' in the image, and provide the bounding boxes (between 0 and 1000, integer) and confidence... The output answer format should be as follows: $<$think$>$ ... $<$/think$>$$<$answer$>$[{'Position': [x1, y1, x2, y2], 'Confidence': number}, ...]$<$/answer$>$ Please strictly follow the format.
\\
\midrule
\textbf{Classification Prompt:} This is an image containing a plant. Please identify the species of the plant based on the image... The output answer format should be as follows: $<$think$>$ ... $<$/think$>$ $<$/think$>$species name$<$/answer$>$ Please strictly follow the format.
\\
\bottomrule
\end{tabular}
}
\vspace{-4mm}
\end{center}
\end{table}

\subsubsection{Data Preparation}\label{sec:data_prepare}
To train \methodname on various visual perception tasks, we need to construct the multimodal training dataset.
Similar to DeepSeek-R1, to enhance the model's reasoning ability and apply this ability to improve visual perception, we designed a prompt format for \methodname to guide the model to output its reasoning process before providing the final answer.
The prompts used for detection and classification tasks are shown in Tab~\ref{tab:prompt}.

During the training process, we use the format reward to guide the model to output its reasoning process and the final answer in a structured format.
The reasoning process is key to the model's self-learning and improvement during reinforcement fine-tuning, while the reward determined by the answer directs the model's optimization.

\begin{figure*}[t]
    \centering
   \includegraphics[width=.98\linewidth]{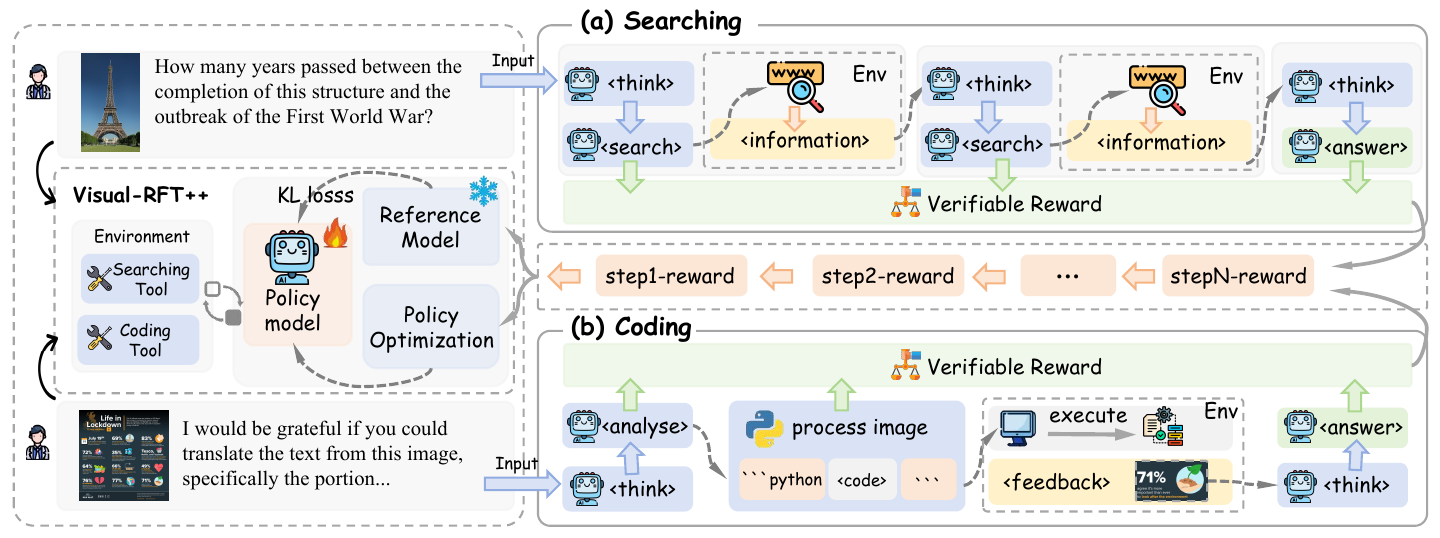}
   \vspace{-8pt}
    \caption{\textbf{Framework of \methodname for Multimodal Agentic Tasks.} We successfully empower LVLMs with multimodal agentic capabilities, including (a) agentic search and (b) agentic coding, enabling them to solve complex multimodal tasks through reasoning, decomposition, and tool interaction.}
    \vspace{-12pt}
    \label{fig:framework_arft}
\end{figure*}

\subsection{\methodname for Multimodal Agentic Tasks}

Fig.~\ref{fig:framework_arft} shows the overall framework of \methodname for multimodal agentic tasks, which is applied to two multimodal agentic scenarios: search and coding.
Given multimodal inputs, the policy model $\pi_\theta$ generates a sequence of responses that include both intermediate reasoning steps (e.g., \texttt{<think>}) and action decisions, such as invoking a search tool or generating Python code to solve complex tasks.

In the search setting, the model is presented with complex multimodal multi-hop VQA queries that often involve multiple sub-questions and require external factual knowledge. Such queries are difficult to resolve using only in-context knowledge or Chain-of-Thought (CoT) prompting. Therefore, the model must be capable of reasoning, decomposing the original question into manageable sub-tasks, and invoking tools to retrieve relevant information from external sources. To address these challenges, \methodname trains the model to autonomously analyze and decompose multi-hop queries and solve them through iterative interactions with a web search engine.

In the agentic coding setting, the input image may suffer from visual degradation or contain redundant content, where only specific regions are relevant to the question. These conditions impose a significant burden on the model’s visual understanding capabilities. \methodname is designed to handle this by training the model to actively generate executable code conditioned on the visual question. The model submits the code to an external environment to preprocess the image (e.g., rotation, cropping), thereby extracting useful information and enabling accurate question answering.

During training, the model typically requires multiple reasoning steps or repeated tool invocations to complete a single task. To guide this process, we design a rule-based verifiable reward that evaluates both the model’s tool usage and final answers. Rewards are assigned at each step and provide learning signals for policy optimization. We will further discuss how to design the verifiable reward for multimodal agentic tasks in Sec. \ref{sec:methoda_reward}.

\subsection{Verifiable Reward in Multimodal Agentic Tasks}\label{sec:methoda_reward}
The design of reward functions plays a critical role in determining the effectiveness of reinforcement learning. The rewards we design for agentic tasks follow a similar structure to those used in visual perception, consisting of a combination of format rewards and accuracy rewards.

\noindent \textbf{Format Reward.} In agentic tasks, we require the model to generate outputs that strictly follow a predefined format. This form of structured reward encourages the model to engage in step-by-step reasoning and to invoke tools as instructed, thereby improving the reliability of tool execution. In both the Coding and Searching tasks, the model’s reasoning process is enclosed within `\texttt{<think>}\texttt{</think>}' tags. When the model needs to invoke a search or code execution tool, it outputs a `\texttt{<search>}\texttt{</search>}' or `\texttt{<code>}\texttt{</code>}' tag, respectively.
The content within each tool tag, either the search query or the generated code, is then extracted and used to interact with the external environment:
\begin{equation}
R_{\text{format}}(o) = \mathbb{I}\left[o \text{ contains valid tags} \right].
\end{equation}
\noindent \textbf{Accuracy Rewards.} The accuracy reward in \methodname is modular. For the final answers in both the searching and coding tasks, we use the F1 score as the reward to evaluate answer quality ($R_{F1}$). F1 rewards offer greater tolerance than exact match, providing smoother and more informative learning signals. This design better captures the fluency and variability of natural language responses, and contributes to more stable reinforcement learning.

For evaluating the effectiveness of search tool usage, we compute the semantic similarity between the model-generated search query and the ground-truth query using a Sentence Transformer. Compared to F1-based rewards, semantic similarity rewards ($R_{sem}$) are more robust to surface-level variations and better reflect whether the model captures the true intent of the retrieval objective. This is especially important for open-ended or paraphrased queries, where lexical overlap may be low but semantic equivalence still holds.

For the coding component, we assign a reward of 1 to all outputs that fall within executable code blocks, without directly supervising the content of the generated code. Instead of evaluating the correctness of the code itself, we allow the model to freely synthesize code based on its own reasoning. This design encourages flexibility and open-ended problem-solving.
In preliminary experiments, we observed that applying strict correctness-based rewards to code generation led the model to converge on a small set of deterministic solutions, which in turn reduced its ability to generalize across diverse visual reasoning scenarios. By removing content-level supervision on code, we promote broader exploration and maintain the agentic nature of decision making.
\begin{equation}
R_{\text{acc}}(q, o) = 
\begin{cases}
R_{\text{F1}}(o_{\text{ans}}, a), & \text{if } o \text{ in (}<\texttt{answer}>\text{)}, \\
R_{\text{sem}}(o_{\text{search}}, s), & \text{if } o \text{ in (}<\texttt{search}>\text{)}, \\
1, & \text{if } o \text{ in (}<\texttt{code}>\text{).}
\end{cases}
\end{equation}
\paragraph{Total Rewards} The total reward used in agentic tasks combines two components: a format reward that encourages the model to follow the required output structure, and an accuracy reward that evaluates the quality of answers and tool usage. The format reward ensures proper tagging for reasoning and tool invocation, while the accuracy reward provides feedback on the final answer and search content. The overall reward function is defined as:
\begin{equation}
R_{\text{total}}(q, o) = R_{\text{format}}(o) + R_{\text{acc}}(q, o).
\end{equation}

\begin{figure*}[t]
    \centering
   \includegraphics[width=.98\linewidth]{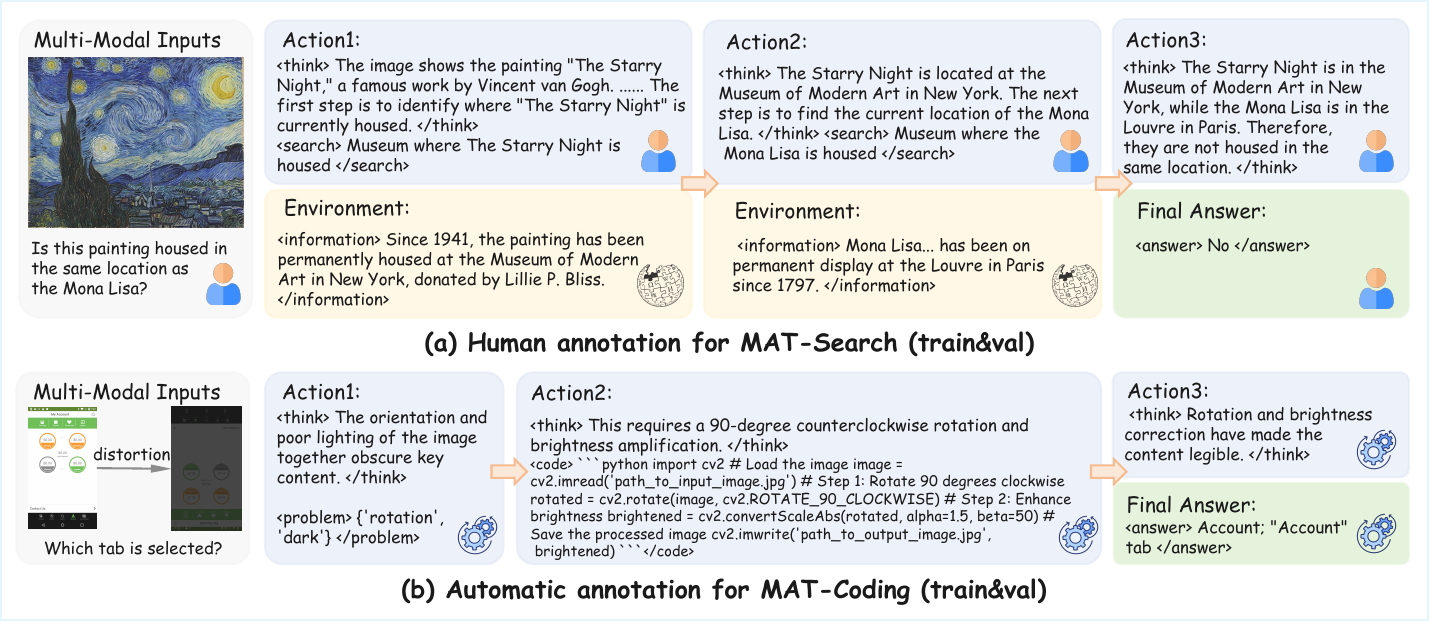}
   \vspace{-8pt}
    \caption{\small \textbf{Data Annotation Pipeline} of our proposed Multimodal Agentic Tool Bench (\benchname): (a) \searchname, a manually annotated and verified dataset for agentic search, and (b) \codename, an automatically generated dataset for agentic coding with a structured pipeline. All contents within the action blocks are generated by the model itself; no distortion-type labels are provided as input at inference time.}
    \vspace{-12pt}
    \label{fig:data_annotation}
\end{figure*}

\noindent \textbf{Multi-Turn Reward.} For multi-turn agentic interactions, we decompose each complete trajectory into multiple trajectory-prefix samples according to the interaction turns. Specifically, a trajectory containing $T$ model responses is converted into $T$ training samples, each retaining the preceding model actions and tool feedback as context and ending at the corresponding response. GRPO is then applied independently to each prefix, and the reward is computed only for its final response. Therefore, rewards are neither accumulated nor discounted across turns. This decomposition enables the model to condition each action on the preceding interaction history while assigning credit only to the action generated at the current turn.

\section{Multimodal Agentic Tool (\benchname) Benchmark} \label{sec:mat}
Given that existing benchmarks for agentic search are predominantly language-only~\cite{bfcl3,ho2020constructing2wiki,yang2018hotpotqa,trivedi2022musique,press2022measuringbamboodle}, and that there is a lack of standardized evaluation protocols for multimodal agentic coding tasks, we construct and annotate a new benchmark to fill this gap.
We refer to the full dataset as the Multimodal Agentic Tool Bench (\benchname), which comprises two sub-settings: \searchname and \codename.
Specifically, we create a set of multimodal multi-hop VQA examples for evaluating agentic searching capabilities, and another set of VQA examples involving image distortions that require tool-based image processing for agentic coding.

\noindent \textbf{\searchname.}
For the agentic search task, as illustrated in Fig.~\ref{fig:data_annotation} (a), we manually construct a multimodal multi-hop VQA dataset. Specifically, we build the \searchname benchmark by annotating 150 high-quality multimodal multi-hop VQA examples. The questions in \searchname vary in difficulty and require different levels of reasoning depth—more complex queries involve more inference steps and factual knowledge, challenging the model's ability to handle composite problems and retrieve relevant external information. Each example is carefully crafted and verified by human annotators to ensure clarity, consistency, and suitability for evaluating agentic reasoning and tool use.

\noindent \textbf{\codename.} For the agentic coding task, as shown in Fig.~\ref{fig:data_annotation} (b), we design an automated data annotation pipeline. Starting from existing VQA examples, we apply various types of image distortions—including rotation, brightness adjustment, blurring, noise injection, and hybrid distortions—to increase the difficulty of visual understanding. Under this setup, directly reasoning over the processed image is challenging for the model.
We annotate 200 test samples to evaluate the model’s ability to complete the image distortion tasks. While the model may attempt to answer questions directly from the processed image, it must overcome various forms of visual noise and irrelevant content. In contrast, advanced agentic systems, such as OpenAI-o3 and our \methodname, can invoke code, manipulate the image, and then utilize the processed result to more effectively complete the VQA task.

\noindent \textbf{\benchname Training Data.}
While OpenAI-o3 demonstrates strong capabilities in agentic search and coding, existing open-source multimodal models perform poorly on such tasks.
Therefore, we construct a multimodal agentic training dataset for \methodname. Given the simplicity and regularity of the search task format, we carefully annotate \textbf{20} multimodal QA examples requiring multi-step reasoning, each paired with a well-defined chain of thought and final answer, which are used to train \methodname in the agentic search setting.
In contrast, the agentic coding task involves more complex workflows, including the model's autonomous generation of code.
To support this, we construct a training set of \textbf{1,200} examples to enable effective learning under this setting.

\noindent \textbf{Data Splits, Difficulty, and Quality Control.}
We divide both benchmarks into Simple and Hard subsets. For MAT-Search, difficulty is determined primarily by the number of retrieval hops and supporting evidence pieces required to derive the answer; for MAT-Coding, Simple examples contain a single distortion, whereas Hard examples combine multiple distortions. The training and evaluation sets described above are constructed independently, with no overlap in questions or images. Two PhD-level experts independently inspected all evaluation examples. For MAT-Search, they verified question clarity, visual grounding, reasoning and retrieval chains, and final answers; for MAT-Coding, they checked source images, distortion labels, required image-processing operations, and QA consistency. Ambiguous, incorrectly labeled, or unreliable examples were revised or removed. Training and evaluation questions were written by different annotators and involve disjoint entity sets.

\section{Experiments}

\subsection{Experimental Setup}
We design separate evaluation protocols for visual perception tasks and multimodal agentic tasks, and conduct extensive experiments across a wide range of benchmarks.

\noindent \textbf{Visual Perception.} 
Our method is adaptable to various visual perception tasks. 
We employ a few-shot learning approach, providing the model with a minimal number of samples for training. For image classification and object detection tasks, we adopt a few-shot setting to evaluate the model's fine-grained discriminative and recognition capability, applying reinforcement learning on limited data. 
Then, for the LISA~\cite{lai2024lisa} dataset focusing on reasoning grounding, which demands strong reasoning abilities, we train the model using \methodname and assess its reasoning and perception performance.
Lastly, for open-vocabulary object detection, we evaluate the model's generalization capability by training the Qwen2-VL-2/7B~\cite{wang2024qwen2vl} using \methodname on a subdivided COCO dataset containing 48 base classes. We then test it on 17 novel classes from COCO and 13 rare classes from LVIS~\cite{gupta2019lvis}. The model's visual perception and reasoning abilities are assessed in an open-vocabulary detection setting.
In our detection experiments, we first prompt the model to check whether the category is present in the image, the model then predicts bounding boxes for categories that exist in the images.

\noindent \textbf{Multimodal Agentic Tasks}
For agentic search, we use the Serper API as the retrieval tool, providing access to Google's search engine. We will release the cached top-$k$ search results used in our experiments, together with the corresponding queries and retrieval configuration (including $k$), through our project repository.
This API-based approach avoids the overhead of building a full RAG system and better reflects how agents interact with external environments in real-world settings. We evaluate our method on three settings: \textbf{(1)} agentic coding on our proposed \codename, \textbf{(2)} agentic search on our proposed \searchname;
\textbf{(3)} generalize to existing multihop QA benchmarks, including 2WikiMultihopQA \cite{ho2020constructing2wiki}, HotpotQA \cite{yang2018hotpotqa}, MuSiQue \cite{trivedi2022musique}, and Bamboogle \cite{press2022measuringbamboodle}.
For all settings, we use F1 and Exact Match (EM) metrics, where F1 captures token-level overlap and EM requires an exact string match with the ground-truth answer.

\begin{figure*}[t]
    \begin{center}
    \includegraphics[width=.96\linewidth]{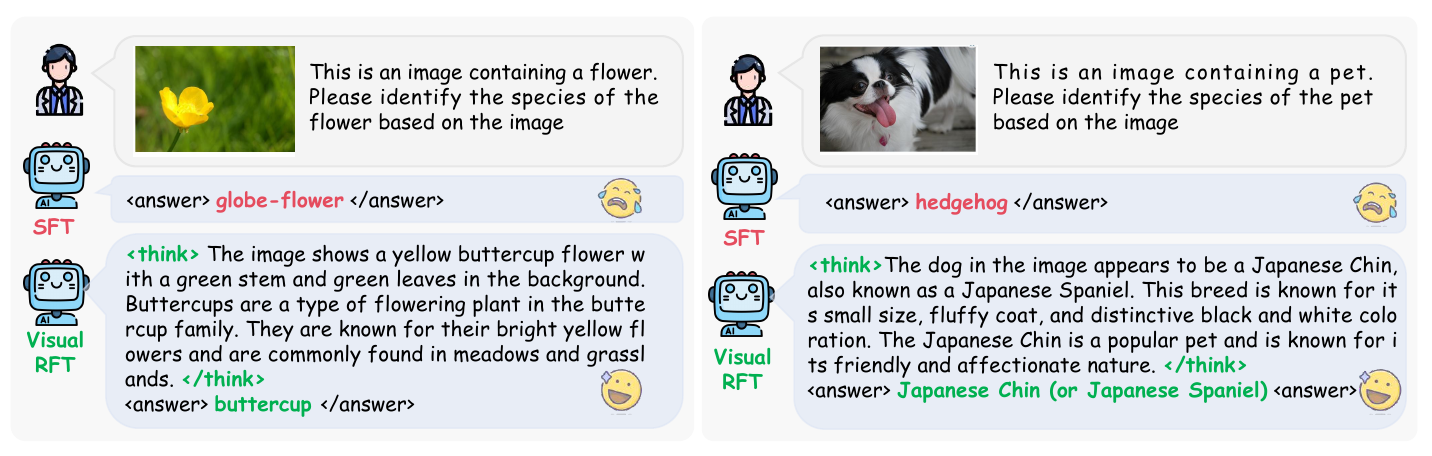}
    \end{center}
    \vspace{-4mm}
    \caption{\small \textbf{Qualitative results of Fine-Grained Image Classification.} The thinking process significantly improves the reasoning ability of LVLMs, leading to higher image classification performance.}
    \label{fig:cls_case}
    \vspace{-4mm}
\end{figure*}

\textbf{Training Details.}
For the multimodal agentic experiments, we use AdamW with a learning rate of $1\times10^{-6}$. Training is performed on 8 GPUs with a per-device batch size of 1 and a gradient accumulation factor of 2, yielding an effective prompt batch size of 16. We sample eight responses per prompt, resulting in 128 generated responses per optimization update. The maximum completion length is 2,048 tokens, and generation uses \texttt{do\_sample=True} with a temperature of 1.0. Training uses BF16 and DeepSpeed ZeRO-3, with parameters and optimizer states offloaded to CPU. We set \texttt{max\_pixels=401408} and the KL coefficient to $\beta=0.04$. For the perception experiments, the IoU threshold is set to $\tau=0.5$. On the same 8-GPU setup, 100 training steps take approximately 76 minutes for MAT-Coding and 60 minutes for MAT-Search.

\subsection{Experiment Results on Visual Perception}

\subsubsection{Few-Shot Classification}
To demonstrate the extensive generalization ability of \methodname in the visual domain, we conduct few-shot experiments on fine-grained image classification. We selected four datasets: Flower102~\cite{flower102}, Pets37~\cite{pets37}, FGVC-Aircraft~\cite{fgvc}, and Car196~\cite{stanfordcars}, which contain dozens to hundreds of similar categories, making the task significantly more difficult to the classification task. 

\begin{table}[t]
\caption{
\small
\textbf{Few-shot results on Fine-grained Classification datasets.} We evaluated four fine-grained image classification datasets under 1-shot,4-shot and 8-shot. Baseline results are from InPK~\cite{Zhou2025InPKIP} under 4-shot setting.}
\vspace{-4mm}
\label{tab:cls}
\begin{center}
\setlength{\tabcolsep}{2.4pt}
\scalebox{0.94}
{
\begin{tabular}{l|c|cccc}
\toprule
\multicolumn{1}{c}{\bf Models}&\multicolumn{1}{c}{\bf Average} & Flower102 & Pets37 & Aircraft & Cars196
\\
\midrule
CoOp~\cite{coop}  & 62.7 & 70.7 & 89.4 & 24.9 & 65.7 \\
MaPLe~\cite{Khattak2022MaPLeMP}  & 67.7 & 80.8 & 92.1 & 29.0 & 68.7 \\
CoCoOp~\cite{cocoop}  & 68.9 & 82.6 & 93.0 & 30.9 & 69.1 \\
PromptSRC~\cite{promptSRC}  & 72.3 & 91.3 & 93.2 & 32.8 & 71.8 \\
\midrule
Qwen2-VL-2B  & 56.0 & 54.8 & 66.4 & 45.9 & 56.8 \\

\midrule 
$+$ SFT (\textit{1-shot}) & 51.7 & 56.6 & 54.7 & 65.3 & 30.0 \\
\rowcolor[HTML]{DAEFF9} $+$ \methodname & 80.3 \footnotesize \hgreen{+24.3} & 70.8 \footnotesize \hgreen{+16.0} & 84.1 \footnotesize \hgreen{+17.7} & 72.5 \footnotesize \hgreen{+26.6} & 93.8 \footnotesize \hgreen{+37.0}\\

\midrule
$+$ SFT (\textit{4-shot})& 55.6 & 58.5 & 55.5 & 67.9 & 40.5 \\
\rowcolor[HTML]{DAEFF9} $+$ \methodname & 81.9 \footnotesize \hgreen{+25.9}& 71.4 \footnotesize \hgreen{+16.6} & 86.1 \footnotesize \hgreen{+19.7} & 74.8 \footnotesize \hgreen{+28.9} & 95.3 \footnotesize \hgreen{+38.5}\\

\midrule
 $+$ SFT(\textit{8-shot})& 60.3 & 59.6 & 71.4 & 69.2 & 40.9 \\
\rowcolor[HTML]{DAEFF9} $+$ \methodname & 85.1 \footnotesize  \hgreen{+29.1} & 77.7 \footnotesize  \hgreen{+22.9}& 90.2 \footnotesize  \hgreen{+23.8} & 75.9 \footnotesize  \hgreen{+30.0} & 96.5 \footnotesize  \hgreen{+39.7}\\

\bottomrule
\end{tabular}
}
\vspace{-6mm}
\end{center}
\end{table}

As shown in Tab.~\ref{tab:cls}, with just one shot of data, \methodname already delivers a significant performance boost (+24.3 percentage points). In contrast, SFT shows a noticeable decline (-4.3 percentage points) with the same minimal amount of data. Under the 4-shot setting, the performance of SFT is still slightly lower than the baseline, while the reinforcement fine-tuned model with \methodname achieves an average performance improvement of 25.9 percentage points. Under the 8-shot settings, as the amount of data increases, SFT's performance slightly exceeds the baseline. However, SFT still lags significantly behind \methodname.
In Fig.\ref{fig:cls_case}, we present some inference cases of the model after reinforcement fine-tuning when handling fine-grained classification tasks. These results demonstrate the strong generalization ability of Visual-RFT++ and its capacity to learn from limited data.

\begin{table}[t]
\caption{
\small
\textbf{Few-Shot results on COCO dataset of 8 categories.} We conducted 1-shot, 4-shot and 8-shot experiments on 8 categories from the COCO dataset.}
\vspace{-4mm}
\label{tab:few-shot_coco}
\begin{center}
\setlength{\tabcolsep}{2.1pt}
\scalebox{.94}
{
\begin{tabular}{l|l|cccccccc}
\toprule
\multicolumn{1}{c}{\bf Models} &\multicolumn{1}{c}{\bf mAP} &\rotatebox{90}{\bf bus} &\rotatebox{90}{\bf train} &\rotatebox{90}{\bf fire hydrant} &\rotatebox{90}{\bf stop sign} &\rotatebox{90}{\bf cat}&\rotatebox{90}{\bf dog}&\rotatebox{90}{\bf bed}&\rotatebox{90}{\bf toilet}
\\
\midrule
Qwen2-VL-2B & 19.6 & 19.0 & 15.8 & 25.8 & 18.4 & 29.9 & 23.2 & 14.6 & 9.8 \\
\midrule
$+$ SFT (\textit{1-shot}) & 19.5 & 18.3 & 17.4 & 23.1 & 18.2 & 28.0 & 23.4 & 17.3 & 10.4 \\
\rowcolor[HTML]{DAEFF9} $+$ \methodname & 33.6 & 23.4 & 35.7 & 39.1 & 23.8 & 54.3 & 42.5 & 19.5 & 30.8 \\
$\Delta$ & \hgreen{+14.0} & \hgreen{+4.4} & \hgreen{+19.9} & \hgreen{+13.3} & \hgreen{+5.4} & \hgreen{+24.4} & \hgreen{+19.3} & \hgreen{+4.9} & \hgreen{+21.0}\\
\midrule
$+$ SFT(\textit{4-shot})  & 25.2 & 25.4 & 23.6 & 26.6 & 21.5 & 35.6 & 30.6 & 18.4 & 19.9 \\
\rowcolor[HTML]{DAEFF9} $+$ \methodname & 40.6 & 30.0 & 40.6 & 45.7 & 35.0 & 60.9 & 44.9 & 24.6 & 43.1 \\
$\Delta$ & \hgreen{+21.0} & \hgreen{+11.0} & \hgreen{+24.8} & \hgreen{+19.9} & \hgreen{+16.6} & \hgreen{+31.0} & \hgreen{+21.7} & \hgreen{+10.0} & \hgreen{+33.3}\\
\midrule
$+$ SFT (\textit{8-shot})& 30.2 & 25.8 & 35.1 & 29.4 & 21.9 & 44.5 & 39.0 & 22.6 & 23.5 \\
\rowcolor[HTML]{DAEFF9} $+$ \methodname & 47.4 & 36.2 & 47.9 & 50.4 & 47.7 & 65.2 & 57.0 & 30.4 & 44.0 \\
$\Delta$ & \hgreen{+27.8} & \hgreen{+17.2} & \hgreen{+32.1} & \hgreen{+24.6} & \hgreen{+29.3} & \hgreen{+35.3}& \hgreen{+33.8} & \hgreen{+15.8} & \hgreen{+34.2}\\

\bottomrule
\end{tabular}
}
\vspace{-2mm}
\end{center}
\end{table}

\begin{table}[htbp]
\caption{
\small
\textbf{Few-shot results on LVIS dataset of 6 rare categories.} We conducted 10-shot experiments on 6 rare categories from the LVIS dataset.}
\vspace{-4mm}
\label{tab:few-shot_lvis}
\begin{center}
\setlength{\tabcolsep}{2pt}
\scalebox{0.94}
{
\begin{tabular}{l|l|cccccc}
\toprule
\multicolumn{1}{c}{\bf Models} &\multicolumn{1}{c}{\bf mAP} & buggy & die & kitchen-table & omelet & papaya & stepladder
\\
\midrule
Qwen2-VL-2B & 4.0 & 2.9 & 1.2 & 13.4 & 4.7 & 1.5 & 0.0\\
\midrule
$+$ SFT & 10.0 & 7.0 & 7.6 & 34.1 & 4.7 & 6.3 & 0.0\\
\rowcolor[HTML]{DAEFF9}  $+$ \methodname & 19.4 & 9.1 & 19.6 & 42.2 & 20.4 & 14.5 & 10.9 \\
$\Delta$ & \hgreen{+15.4} & \hgreen{+6.2} & \hgreen{+18.4} & \hgreen{+29.2} & \hgreen{+15.7}& \hgreen{+13.0}& \hgreen{+10.9}\\
\midrule
Qwen2-VL-7B & 15.4 & 19.7 & 21.9 & 14.5 & 18.1 & 18.5 & 0.0 \\
\midrule
$+$ SFT & 27.6 & 26.9 & 21.9 & 49.7 & 29.2 & 25.2 & 12.7 \\
\rowcolor[HTML]{DAEFF9}  $+$ \methodname & 33.8 & 26.2 & 27.8 & 70.6 & 23.5 & 21.2 & 29.3 \\
$\Delta$ & \hgreen{+18.4} & \hgreen{+6.5} & \hgreen{+5.9} & \hgreen{+56.1} & \hgreen{+5.4}& \hgreen{+2.7}& \hgreen{+29.3}\\
\bottomrule
\end{tabular}
}
\vspace{-2mm}
\end{center}
\end{table}

\subsubsection{Few-Shot Object Detection}
Few-shot learning challenges traditional visual-language models and large-scale vision-language models (LVLMs). Reinforcement fine-tuning provides a new solution to this problem by enabling the model to quickly learn and understand with a small amount of data. We selected eight categories from the COCO dataset, with 1, 4 and 8 images per category, to construct training sets with limited data. For the LVIS dataset, we select 6 rare categories. Since the training images for these rare categories are very sparse, with each category having between 1 and 10 images, we approximated this as a 10-shot setting. We then train the Qwen2-VL-2/7B model for 200 steps using both reinforcement fine-tuning and SFT, to evaluate the model’s learning ability with limited data.

As shown in Tab.~\ref{tab:few-shot_coco} and Tab.~\ref{tab:few-shot_lvis}, although both SFT and reinforcement fine-tuning can improve the model's recognition accuracy under the few-shot setting, the model after reinforcement fine-tuning consistently outperforms the SFT model by a large margin, maintaining a significant lead. On the COCO~\cite{coco} categories, as the training data increases, the SFT model reaches an average mAP of approximately $30$, while the reinforcement fine-tuned model approaches $47$. In the LVIS~\cite{gupta2019lvis} few-shot experimental results shown in Tab.~\ref{tab:few-shot_lvis}, for the six more challenging rare categories in LVIS, reinforcement fine-tuning still outperforms SFT. The results in Tab.~\ref{tab:few-shot_coco} and Tab.~\ref{tab:few-shot_lvis} clearly demonstrate the exceptional performance of reinforcement fine-tuning in the few-shot setting, where the model achieves a significant improvement in visual perception capabilities through reinforcement learning with only a small amount of data.

\begin{figure*}[t]
    \begin{center}
    \includegraphics[width=.96\linewidth]{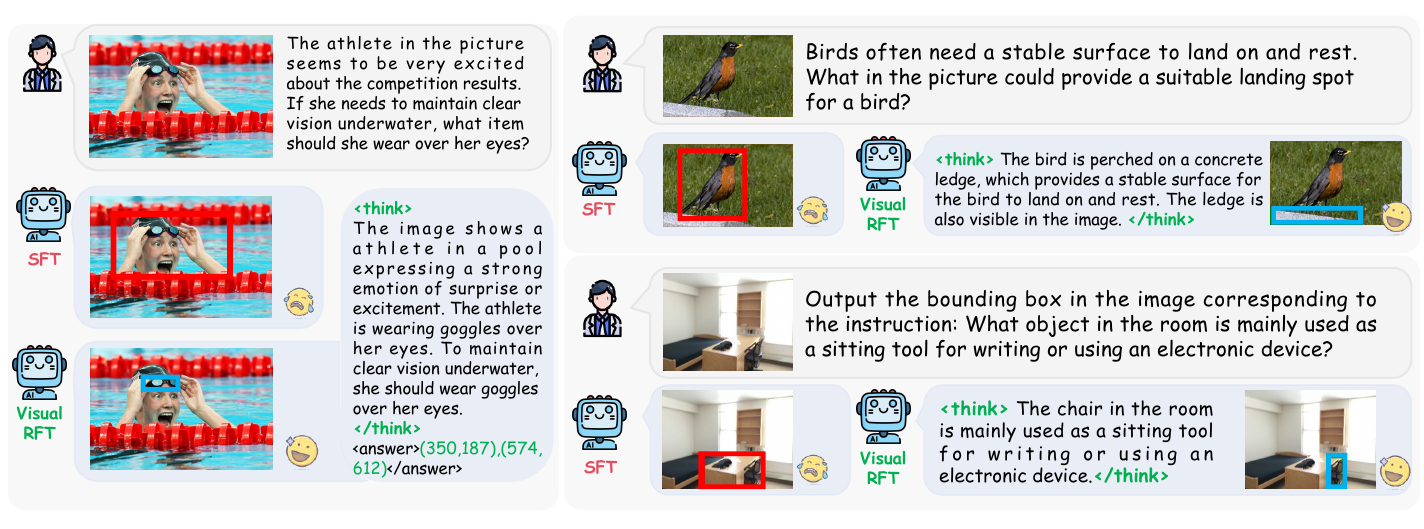}
    \end{center}
    \vspace{-4mm}
    \caption{\small \textbf{Qualitative results of reasoning grounding on LISA~\cite{lai2024lisa}.} Thinking process significantly improves reasoning grounding ability with \methodname.}
    \label{fig:lisa_case}
    \vspace{-2mm}
\end{figure*}

\subsubsection{Reasoning Grounding}
Another crucial aspect of vision-language intelligence is grounding the exact object according to user needs. Previous specialized detection systems lack reasoning abilities and fail to fully understand the user's intentions. Pioneered by LISA~\cite{lai2024lisa}, previous works enable large language models (LLMs) to output control tokens for other models (such as SAM \cite{kirillov2023segment}) or directly predict bounding box coordinates \cite{wang2024qwen2vl,rasheed2024glamm} through supervised fine-tuning. In our work, we explore the use of \methodname in this task and find that reinforcement learning (RL) leads to significant improvements over supervised fine-tuning.
\begin{table}[t]
    \caption{\textbf{Reasoning Grounding Results on LISA~\cite{lai2024lisa}}. 
    \methodname surpasses SFT in reasoning grounding with 239 training images.}
    \vspace{-6pt}
  \centering
  \setlength{\tabcolsep}{10pt}
  \scalebox{1.}
    {
    \begin{tabular}{l|ccc}
    \toprule
    Model  & mIoU$_{\text{test}}$ & mIoU$_{\text{val}}$ & gIoU$_{\text{test}}$\\
    \midrule
    OV-Seg~\cite{liang2023open} & 28.4 & 30.5 & 26.1 \\
    X-Decoder~\cite{zou2023generalized} & 28.5 & 29.1 &  24.3 \\
    GroundedSAM~\cite{liu2024grounding} & 26.2 & 28.6 & 21.3 \\
    \midrule
    Qwen2-VL-2B & 26.9 & 30.1 & 25.3 \\
    \midrule
    $+$ SFT   & 28.3 & 29.7 & 25.3 \\
    \rowcolor[HTML]{DAEFF9} $+$ \methodname   & 37.6 \footnotesize \hgreen{+10.7} & 34.4 \footnotesize \hgreen{+4.3} & 34.4 \footnotesize \hgreen{+9.1}\\
    \midrule
    Qwen2-VL-7B & 40.4 & 45.2 & 38.0 \\
    \midrule
    $+$ SFT   & 39.1  & 43.9 & 37.2 \\
    \rowcolor[HTML]{DAEFF9} $+$ \methodname  & 43.9 \footnotesize \hgreen{+3.5} & 47.1 \footnotesize \hgreen{+1.9} & 43.7 \footnotesize \hgreen{+5.6}\\
    \bottomrule
    \end{tabular}%
    }
  \label{tab:lisa_grounding}%
\vspace{-4pt}
\end{table}%

\begin{table}[t]
\caption{
\small
\textbf{Open Vocabulary Object Detection Results on COCO dataset.} We trained on 48 base categories for 10-shot and tested on 17 novel categories.}
\vspace{-4mm}
\label{tab:coco_ov}
\begin{center}
\setlength{\tabcolsep}{10pt}
\scalebox{1.0}
{
\begin{tabular}{l|llll}
\toprule
\multicolumn{1}{c}{\bf Models} &\multicolumn{1}{c}{\bf $nAP$}&\multicolumn{1}{c}{\bf $bAP$}&\multicolumn{1}{c}{\bf $mAP$}
\\
\midrule
Qwen2-VL-2B & 9.6 & 5.6 & 6.7 \\
\midrule
$+$ SFT & 16.7 & 12.4 & 13.5 \\
\rowcolor[HTML]{DAEFF9} $+$ \methodname & 27.9 \footnotesize \hgreen{+18.3}& 20.3 \footnotesize \hgreen{+14.7}& 22.2 \footnotesize \hgreen{+15.5}\\
\midrule
Qwen2-VL-7B & 25.0 & 17.6 & 19.5 \\
\midrule
$+$ SFT & 26.9 & 18.4 & 20.6 \\
\rowcolor[HTML]{DAEFF9} $+$ \methodname & 32.3 \footnotesize \hgreen{+7.3}& 23.6 \footnotesize \hgreen{+6.0}& 25.9 \footnotesize \hgreen{+6.4}\\
\bottomrule
\end{tabular}
}
\end{center}
\vspace{-4mm}
\end{table}

We fine-tune the Qwen2-VL-2B/7B models \cite{wang2024qwen2vl} using \methodname and supervised fine-tuning (SFT) on the LISA training set, which consists of 239 images with reasoning grounding objects. We follow the same test setting with LISA and compare the results of SFT and our method, both with 500 fine-tuning steps. As shown in Tab.~\ref{tab:lisa_grounding}, \methodname significantly improves the final results in terms of bounding box IoU compared to SFT. Additionally, we prompt SAM~\cite{kirillov2023segment} with the Qwen2-VL predicted bounding box to generate the segmentation mask (evaluated using gIoU). \methodname significantly enhances grounding ability and outperforms previous specialized detection systems. Qualitative results are visualized in Fig.~\ref{fig:lisa_case}, where the thinking process significantly improves the ability to reason and grounding accuracy. Through \methodname, Qwen2-VL learns to think critically and carefully examine the image to produce accurate grounding results.

\begin{figure}[t]
    \vspace{-2mm}
    \begin{center}
    \includegraphics[width=1.\linewidth]{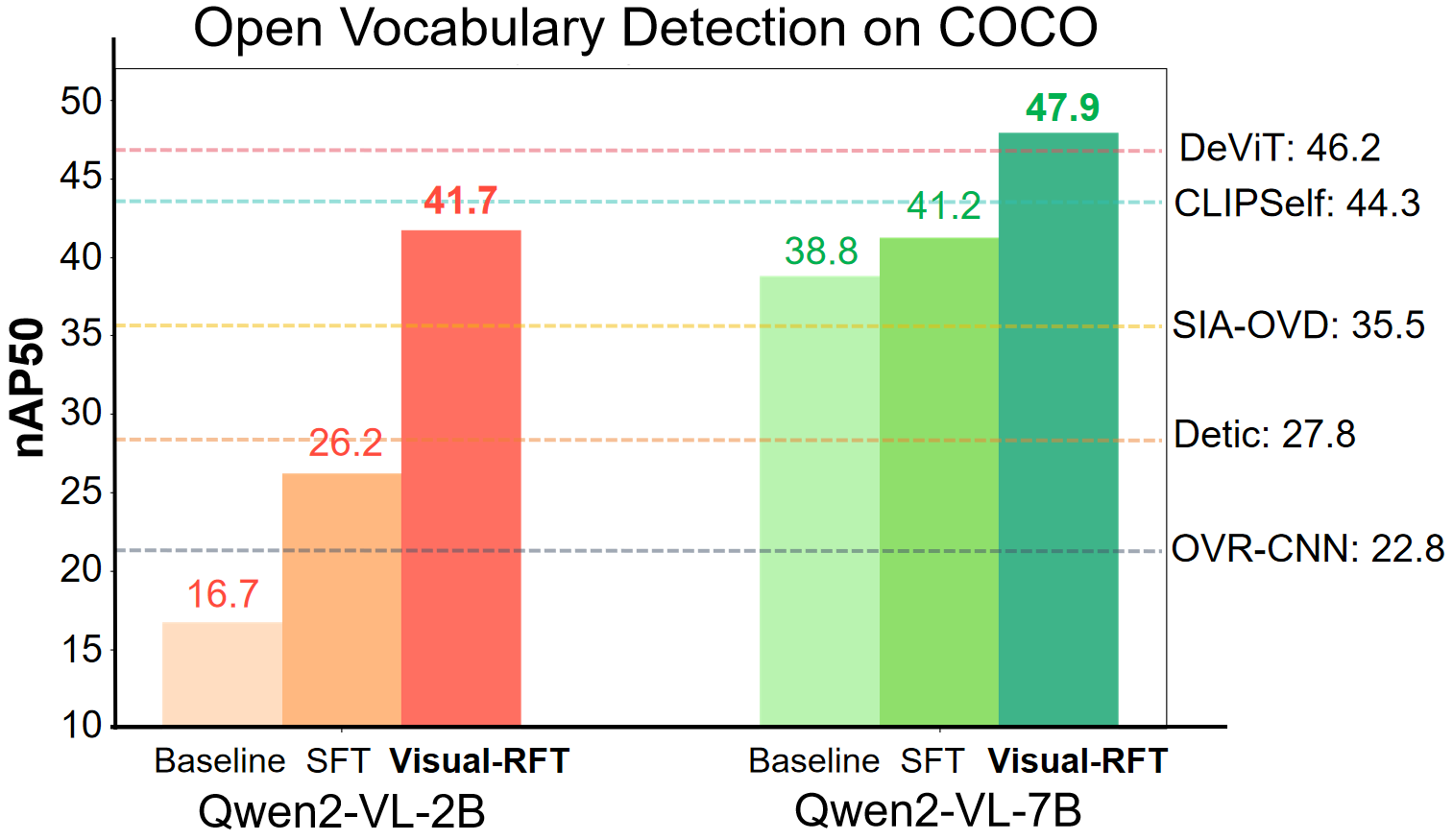}
    \end{center}
    \vspace{-2mm}
    \caption{\small \textbf{Open Vocabulary Results Compared with Specialized Classification Models.} We display the performance of different models in the figure, using nAP50 as the evaluation metric.}
    \label{fig:sota}
    \vspace{-2mm}
\end{figure}

\begin{table*}[t]
\caption{
\small
\textbf{Results of \benchname.} We conducted experiments on \benchname, including \codename and \searchname, and the table presents the evaluation results of several open-source and proprietary models on our benchmark. We use F1 score and Exact Match (EM) to evaluate model performance.}
\vspace{-10pt}
\label{tab:coding_bench}
\begin{center}
\setlength{\tabcolsep}{4.7pt}
\renewcommand{\arraystretch}{1.1}
\scalebox{1.}{
\begin{tabular}{l|c|
cc|cc|
cc|cc|cc|cc}
\toprule
\multirow{3}{*}{\textbf{Models}} & 
\multirow{3}{*}{\shortstack{\textbf{Reasoning} \\ \textbf{with} \\ \textbf{Tools} }} &
\multicolumn{6}{c|}{\textbf{\codename}} &
\multicolumn{6}{c}{\textbf{\searchname}} \\
\cmidrule(lr){3-8} \cmidrule(lr){9-14}
& &
\multicolumn{2}{c|}{\textbf{Simple}} &
\multicolumn{2}{c|}{\textbf{Hard}} &
\multicolumn{2}{c|}{\textbf{Avg}} &
\multicolumn{2}{c|}{\textbf{Simple}} &
\multicolumn{2}{c|}{\textbf{Hard}} &
\multicolumn{2}{c}{\textbf{Avg}} \\
\cmidrule(lr){3-4} \cmidrule(lr){5-6} \cmidrule(lr){7-8} \cmidrule(lr){9-10} \cmidrule(lr){11-12} \cmidrule(lr){13-14}
& &
\textit{F1} & \textit{EM} 
& \textit{F1} & \textit{EM}
& \textit{F1} & \textit{EM}
& \textit{F1} & \textit{EM}
& \textit{F1} & \textit{EM}
& \textit{F1} & \textit{EM}\\
\midrule
GPT-4o~\cite{hurst2024gpt4o}  & \xmark  & 47.12 & 38.57 & 27.57 & 15.38 & 34.41 & 23.5 & 68.55 & 61.33 & 53.61 & 42.67 & 61.08 & 52.00 \\
OpenAI-o3~\cite{OpenAI-o3}  & \checkmark  & 70.38 & 65.38 & 75.00 & 70.59 & 72.99 & 68.33 & 79.72 & 70.67 & 63.74 & 52.00 & 71.73 & 61.33\\
\midrule
LLaVa-v1.5-7B~\cite{llava1.5}  & \xmark & 19.50 & 12.86 & 9.30 & 5.38 & 12.87 & 8.00 & 56.55 & 52.00 & 30.32 & 25.33 & 43.44 & 38.67 \\
LLaVa-Next-7B~\cite{li2024llavanext}  & \xmark & 30.78 & 17.14 & 17.11 & 10.00 & 21.89 & 12.5 & 63.27 & 56.00 & 38.75 & 29.33 & 51.01 & 42.67 \\
LLaVa-OneVision-7B~\cite{li2024llavaov}  & \xmark & 39.86 & 28.57 & 16.05 & 11.54 & 24.38 & 17.5 &61.78 & 54.67 & 31.66 & 26.67 & 46.72 & 40.67 \\
Xcomposer2.5~\cite{xcomposer2.5}  & \xmark & 36.06 & 22.86 & 19.90 & 10.77 & 25.56 & 15.0 & 60.16 & 54.67 & 31.93 & 28.00 & 46.04 & 41.33 \\
InternVL2.5-8B~\cite{internvl2.5}  & \xmark & 39.48 & 28.57 & 26.62 & 13.85 & 31.12 & 19.00 & 61.72 & 53.33 & 41.69 & 33.33 &  51.70 & 43.33 \\
\midrule
Qwen2.5-VL-3B~\cite{bai2025qwen25vl}  & \xmark & 46.29 & 35.71 & 17.98 & 13.85 & 27.89 & 21.50 & 57.54 & 50.67 & 33.11 & 26.67 & 45.32 & 38.67 \\
\rowcolor[HTML]{DAEFF9} \quad + \methodname & \checkmark & 49.78 & 40.00 & 28.42 & 13.08 & 35.90 & 22.50 & 56.41 & 50.67 & 45.55 & 36.00 & 50.98 & 43.33 \\
\quad $\Delta$ & -- & \hgreen{+3.49} & \hgreen{+4.29} & \hgreen{+10.44} & \hblue{-0.78} & \hgreen{+8.01} & \hgreen{+1.0} & \hblue{-1.13} & \hgreen{+0.0} & \hgreen{+12.44} & \hgreen{+9.33} & \hgreen{+5.66} & \hgreen{+4.66} \\
\midrule
Qwen2.5-VL-7B~\cite{bai2025qwen25vl}  & \xmark & 55.23 & 40.00 & 19.67 & 11.54 & 32.12 & 21.50 & 67.40 & 61.33 & 39.59 & 32.00 & 53.49 & 46.67 \\
\rowcolor[HTML]{DAEFF9} \quad + \methodname & \checkmark & 60.10 & 51.43 & 45.60 & 25.38 & 50.68 & 34.50 & 71.78 & 66.67 & 55.77 & 44.00 & 63.77 & 55.33 \\
\quad $\Delta$ & -- & \hgreen{+4.87} & \hgreen{+11.43} & \hgreen{+25.93} & \hgreen{+13.84} & \hgreen{+18.56} & \hgreen{+13.00} & \hgreen{+4.38} & \hgreen{+5.37} & \hgreen{+16.18} & \hgreen{+12.00} & \hgreen{+10.28} & \hgreen{+8.66} \\
\bottomrule
\end{tabular}
}
\end{center}
\vspace{-12pt}
\end{table*}

\begin{figure*}[t]
    \centering
   \includegraphics[width=.98\linewidth]{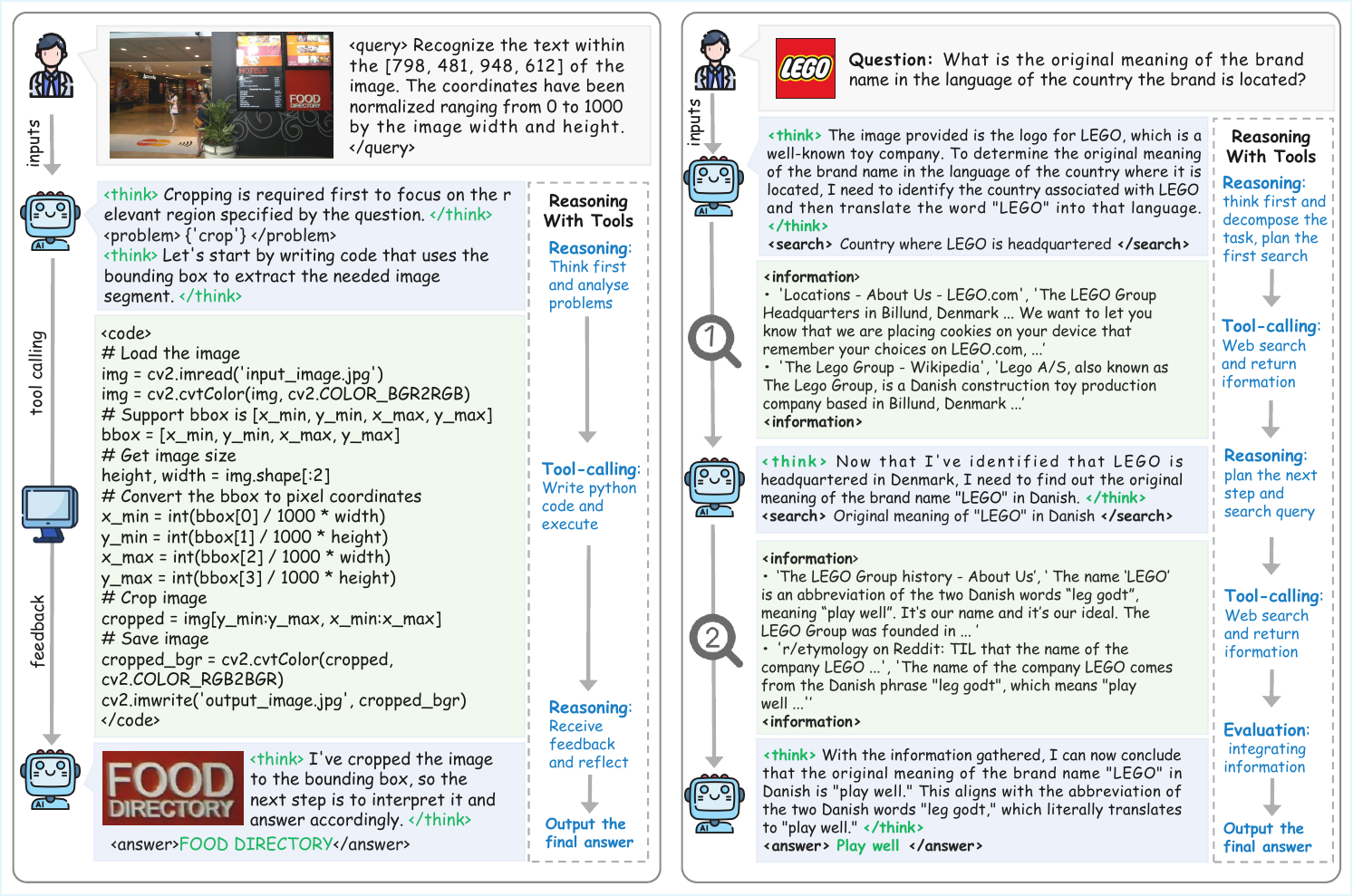}
   \vspace{-4pt}
    \caption{\small \textbf{Visualization Inference Cases of \methodname.} Demonstrating \methodname's multimodal agentic capabilities: processing an image and answering a question via code generation and execution (\textbf{left}), and solving multi-hop VQA through query decomposition and search tool invocation (\textbf{right}).
}
    \label{fig:case_mat_infere}
    \vspace{-16pt}
\end{figure*}

\subsubsection{Open Vocabulary Object Detection}
To further demonstrate the powerful generalization ability of reinforcement fine-tuning, we conduct open vocabulary object detection experiments. We constructed a 10-shot training set by selecting 10 images per category from the 48 base categories of the COCO dataset. We perform \methodname and SFT on the Qwen2-VL-2/7B model~\cite{wang2024qwen2vl} using this data, and test the model on 17 new categories it has never seen before. 

As shown in Tab.~\ref{tab:coco_ov}, after reinforcement fine-tuning, the Qwen2-VL-2/7B model achieves an average nAP increase of $18.3$ and $7.3$ on 17 new categories from the COCO dataset. The \methodname transfers its detection capabilities from the COCO base categories to the new COCO categories. In Fig.~\ref{fig:sota}, we compare \methodname's performance with several other detection models~\cite{Zhang2023DetectET,Wu2023CLIPSelfVT, Wang2024SIAOVDSA,detic,ovrcnn} using nAP50 as the evaluation metric. Both Qwen2-VL-2B and Qwen2-VL-7B models show significant improvements, with the 7B model outperforming the specialized model DE-ViT~\cite{Zhang2023DetectET} under 10-shot setting. This demonstrates that \methodname has a significant impact on improving the performance and generalization ability in visual recognition for LVLMs. 

\subsection{Experiment Results on Multimodal Agentic Tasks}

\subsubsection{Results on \codename}
We evaluate our method on the \codename benchmark. Compared to the base Qwen2.5-VL-3B/7B models, \methodname enables models to autonomously analyze distorted or misaligned images—including both degraded and crop-requiring inputs—generate preprocessing code, and complete the QA pipeline, showcasing their agentic capabilities in reasoning, task decomposition, and tool use.

As shown in Tab.\ref{tab:coding_bench}, the Qwen2.5-VL-7B model achieves substantial improvements with \methodname, gaining +18.56 F1 and +13.00 EM on average. The gains are especially notable on the Hard subset, where base models struggle due to limited reasoning and no tool-use ability. With \methodname, the model invokes code-based tools (e.g., repair, crop), resulting in strong performance gains. Example cases are shown in Fig.\ref{fig:case_mat_infere}. The 3B model also benefits from \methodname, with consistent F1 improvements. EM remains stable due to strict formatting requirements, but the F1 gains reflect improved task completion ability enabled by agentic reasoning with tool.

\noindent \textbf{Statistical Reliability.} Given the compact MAT test sets, we report
95\% bootstrap confidence intervals and paired significance tests. On MAT-Search,
Visual-RFT++ achieves 63.77 F1 (95\% CI $[61.9, 65.4]$), significantly outperforming
GPT-4o (61.08 F1, CI $[59.3, 62.8]$; paired bootstrap $p<0.05$), while OpenAI-o3
remains significantly higher (71.73 F1, $p<0.001$).

We also observe that OpenAI-o3, with native tool-use ability, achieves the highest performance, surpassing GPT-4o. Its ability to reason and manipulate images via code highlights the strength of agentic multimodal models. In summary, \methodname significantly improves model performance on \codename, especially for harder tasks. By introducing agentic capabilities, it enables LVLMs to reason, invoke tools, and solve complex visual problems effectively.

\begin{table*}[t]
\caption{
\small \textbf{Results on Existing Text-Only Multihop QA Benchmarks.} Our \methodname outperforms strong baselines such as Search-o1 \cite{li2025search-o1}, Search-R1 \cite{jin2025search-r1}, and ZeroSearch \cite{sun2025zerosearch}.}
\label{tab:multihopqa}
\vspace{-8pt}
\begin{center}
\setlength{\tabcolsep}{7.7pt}
\renewcommand{\arraystretch}{1.1}
\scalebox{1.}{
\begin{tabular}{l|c|
cc|cc|cc|cc|cc}
\toprule
\multirow{2}{*}{\textbf{Models}} & 
\multirow{2}{*}{\shortstack{\textbf{Reasoning} \\ \textbf{with Tools}}} &
\multicolumn{2}{c|}{\textbf{2Wiki} \cite{ho2020constructing2wiki}} &
\multicolumn{2}{c|}{\textbf{HotpotQA} \cite{yang2018hotpotqa}} &
\multicolumn{2}{c|}{\textbf{MuSiQue} \cite{trivedi2022musique}} &
\multicolumn{2}{c|}{\textbf{Bamboogle} \cite{press2022measuringbamboodle}} &
\multicolumn{2}{c}{\textbf{Avg}} \\
\cmidrule(lr){3-4} \cmidrule(lr){5-6} \cmidrule(lr){7-8} \cmidrule(lr){9-10} \cmidrule(lr){11-12}
& &
\textit{F1} & \textit{EM} 
& \textit{F1} & \textit{EM}
& \textit{F1} & \textit{EM}
& \textit{F1} & \textit{EM}
& \textit{F1} & \textit{EM}\\
\midrule
\multicolumn{10}{l}{\textbf{\textit{Qwen-2.5-7B-Instruct}}~\cite{bai2025qwen25vl}} \\
\quad + Direct Inference & \xmark  & -- & 25.00 & -- & 18.30 & -- & 3.10 & -- & 12.00 & -- & 14.60\\
\quad + CoT & \xmark  & -- & 11.00 & -- & 9.20 & -- & 2.20 & -- & 23.20 & -- & 11.40\\
\quad + IRCoT & \xmark  & -- & 14.90 & -- & 13.30 & -- & 7.20 & -- & 22.40 & -- & 14.45\\
\quad + RAG & \xmark  & -- & 23.50 & -- & 29.90 & -- & 5.80 & -- & 20.80 & -- & 20.00\\
\midrule
\quad + Search-o1~\cite{li2025search-o1} & \checkmark  & -- & 17.60 & -- & 18.70 & -- & 5.80 & -- & 29.60 & -- & 17.93\\
\quad + Search-R1~\cite{jin2025search-r1}  & \checkmark  & -- & 41.40 & -- & \textbf{37.00} & -- & 14.60 & -- & 36.80 & -- & 32.45\\
\quad + ZeroSearch~\cite{sun2025zerosearch}  & \checkmark  & -- & 43.12 & -- & 29.21 & -- & \textbf{19.72} & -- & 35.20 & -- & 31.81\\
\midrule
\multicolumn{10}{l}{\textbf{\textit{Qwen2.5-VL-3B-Instruct}}~\cite{bai2025qwen25vl}} \\
\quad + Direct Inference  & \xmark & 31.60 & 26.40 & 22.90 & 15.84  & 9.64 & 2.36 & 10.82 & 5.60 & 18.74 & 12.55 \\
\quad + RAG  & \xmark & 35.14 & 26.92 & 35.08 & 23.67 & 15.08 & 7.61 & 28.37 & 19.20 & 28.42 & 19.35 \\
									
\rowcolor[HTML]{DAEFF9} \quad + \methodname & \checkmark & 49.77 & 39.13 & 41.33 & 30.10 & 19.10 & 11.58 & 49.01 & 38.40 & 39.80 & 29.80 \\
\quad $\Delta$ & -- & \hgreen{+18.17} & \hgreen{+12.73} & \hgreen{+18.43} & \hgreen{+14.26} & \hgreen{+9.46} & \hgreen{+9.22} & \hgreen{+38.19} & \hgreen{+32.80} & \hgreen{+21.06} & \hgreen{+17.25} \\
\midrule
\multicolumn{10}{l}{\textbf{\textit{Qwen2.5-VL-7B-Instruct}}~\cite{bai2025qwen25vl}} \\
\quad + Direct Inference  & \xmark & 27.90 & 22.15 & 25.35 & 17.35 & 9.45 & 2.61 & 14.07 & 5.60 & 19.19 & 11.93 \\
\quad + RAG  & \xmark & 31.58 & 21.74 & 36.33 & 24.09 & 15.36 & 7.61 & 27.60 & 19.20 & 27.72 & 18.16\\
\rowcolor[HTML]{DAEFF9} \quad + \methodname & \checkmark & \textbf{63.09} & \textbf{51.99} & \textbf{48.00} & 36.48 & \textbf{22.71} & 14.07 & \textbf{60.15} & \textbf{48.80} & \textbf{48.49} & \textbf{37.84} \\
\quad $\Delta$ & -- & \hgreen{+35.19} & \hgreen{+29.84} & \hgreen{+22.65} & \hgreen{+19.13} & \hgreen{+13.26} & \hgreen{+11.46} & \hgreen{+46.08} & \hgreen{+43.20} & \hgreen{+29.30} & \hgreen{+25.91} \\
\bottomrule
\end{tabular}
}
\end{center}
\vspace{-16pt}
\end{table*}

\subsubsection{Results on \searchname}
We compare our proposed method against several open-source and proprietary models on \searchname. As shown in Tab.~\ref{tab:coding_bench}, \methodname improves the average F1 score by 5.66 percentage points and the EM score by 4.66 percentage points on Qwen2.5-VL-3B. On Qwen2.5-VL-7B, the improvements are even more significant, with F1 and EM gains of 10.28 percentage points and 8.66 percentage points, respectively.

These gains highlight the effectiveness of reinforcement fine-tuning with verifiable rewards in equipping the model with structured multimodal reasoning and tool-use capabilities. \methodname enables step-by-step planning and dynamic evidence acquisition, which are crucial for solving complex multi-hop queries. Additionally, we include comparisons with traditional approaches such as RAG and CoT in our ablation study (\ref{sec:ablation_arft}) to further validate the benefits of our method.

On MAT-Search, Visual-RFT++ with Qwen2.5-VL-7B exceeds direct-inference GPT-4o without external tools. In the controlled comparison in Tab.~\ref{tab:tool_enabled_baselines}, GPT-4o with the same search tool achieves higher scores, while \methodname improves over the same-backbone ReAct baseline by 8.04 F1 and 7.33 EM percentage points. Similar to \codename, we also note that OpenAI's o3, with its inherently strong reasoning ability, still demonstrates superior performance on \searchname and outperforms all open-source models. This underscores the need for further research into the agentic capabilities of open-source multimodal models.

\subsubsection{Results on Existing Multi-Hop QA Benchmarks}
To further evaluate the generalization ability of \methodname, we conducted comprehensive experiments on several existing text-only multi-hop QA benchmarks, including 2WikiMultihopQA \cite{ho2020constructing2wiki}, HotpotQA \cite{yang2018hotpotqa}, MuSiQue \cite{trivedi2022musique}, and Bamboogle \cite{press2022measuringbamboodle}. 

Here we apply \methodname to Qwen2.5-VL-3B/-7B models using 20 manually annotated multimodal multi-hop VQA training examples (see Sec. \ref{sec:mat}).
Since our training is performed on VQA-style data involving both visual and textual modalities, there exists a clear modality and input-type gap between our training set and these language-only QA benchmarks \cite{ho2020constructing2wiki,yang2018hotpotqa,trivedi2022musique,press2022measuringbamboodle}. As such, the \textbf{out-of-domain} evaluations provide a strong testbed for validating the cross-modal generalization capability of \methodname.

Tab.~\ref{tab:multihopqa} presents a comparison with various baselines built on Qwen2.5-3B/7B, covering traditional approaches such as Chain-of-Thought (CoT) and Retrieval-Augmented Generation (RAG), as well as reinforcement learning-based methods like Search-R1 and ZeroSearch. As shown in Tab.~\ref{tab:multihopqa}, \methodname yields substantial performance gains on these out-of-domain multi-hop QA benchmarks. The Qwen2.5-VL-3B model improves by 21.06 F1 and 17.25 EM percentage points over the baseline. The 7B model further surpasses all baselines, with average EM gains of 25.91 percentage points over direct inference. These improvements stem from \methodname’s ability to perform task decomposition and dynamic tool use, enabling step-by-step reasoning with flexible interaction with external information. Despite using only 20 training examples, \methodname achieves superior data efficiency and strong generalization across unseen domains. Moreover, \methodname is built on LVLMs and outperforms several LMM-based methods, showcasing strong modality transfer and task generalization capabilities.

\subsubsection{Analysis}\label{sec:C4}
Beyond final-answer performance, we further examine whether Visual-RFT++ learns effective tool-use behaviors and their practical overhead in both agentic settings. Specifically, we analyze query formulation, search patterns, and failure modes on MAT-Search, as well as code executability, operation consistency, and the effects of image manipulation on MAT-Coding.

\textbf{Search Behavior Diagnostics.}
To further examine whether these gains reflect improved tool use, we analyze the model's search behavior. During training, the semantic similarity between generated and annotated target queries reaches 0.95 and 0.92 for the 7B and 3B models, respectively. Among 100 manually inspected test trajectories, 86\% of the generated queries are relevant to the current reasoning sub-goal, while only 13.2\% of the search actions are invalid or redundant. The number of tool calls also adapts to task difficulty: Simple questions typically require 2--3 searches, whereas Hard questions require 3--5. Remaining failures mainly arise from incorrect initial visual recognition, incomplete evidence collection, or repeated query reformulation in long reasoning chains.

\textbf{Coding Behavior Diagnostics.}
Beyond final-answer accuracy, we examine whether the generated programs are executable and perform the intended image operations. Overall, 98.5\% and 96.5\% of the expected coding steps from the 7B and 3B models produce executable code, respectively, indicating that basic executability is rarely the primary bottleneck. For the 7B model, 86.34\% of the generated programs match the intended operation family, with higher rates for \textit{dark} (94.6\%) and \textit{overexposure} (90.9\%) than for \textit{rotation180} (68.2\%), where incorrect rotation angles or directions remain common. As illustrated in Fig.~9, appropriate cropping or restoration can expose question-relevant content, whereas incorrect operations may remove useful evidence and hurt the final answer. Thus, executable code alone is insufficient; accurate visual diagnosis and operation selection are also essential for effective tool use.

\textbf{Efficiency Analysis.}
Tool use introduces additional inference overhead compared with direct prediction. Across MAT, tool-enabled inference requires $2.79\times$ as many tokens and incurs $1.53\times$ the average inference latency. Together with the tool-call frequencies reported above, these statistics characterize the practical trade-off between performance improvements and inference cost.

\subsection{Ablation Study}

\begin{figure}[t]
    \begin{center}
    \includegraphics[width=1.\linewidth]{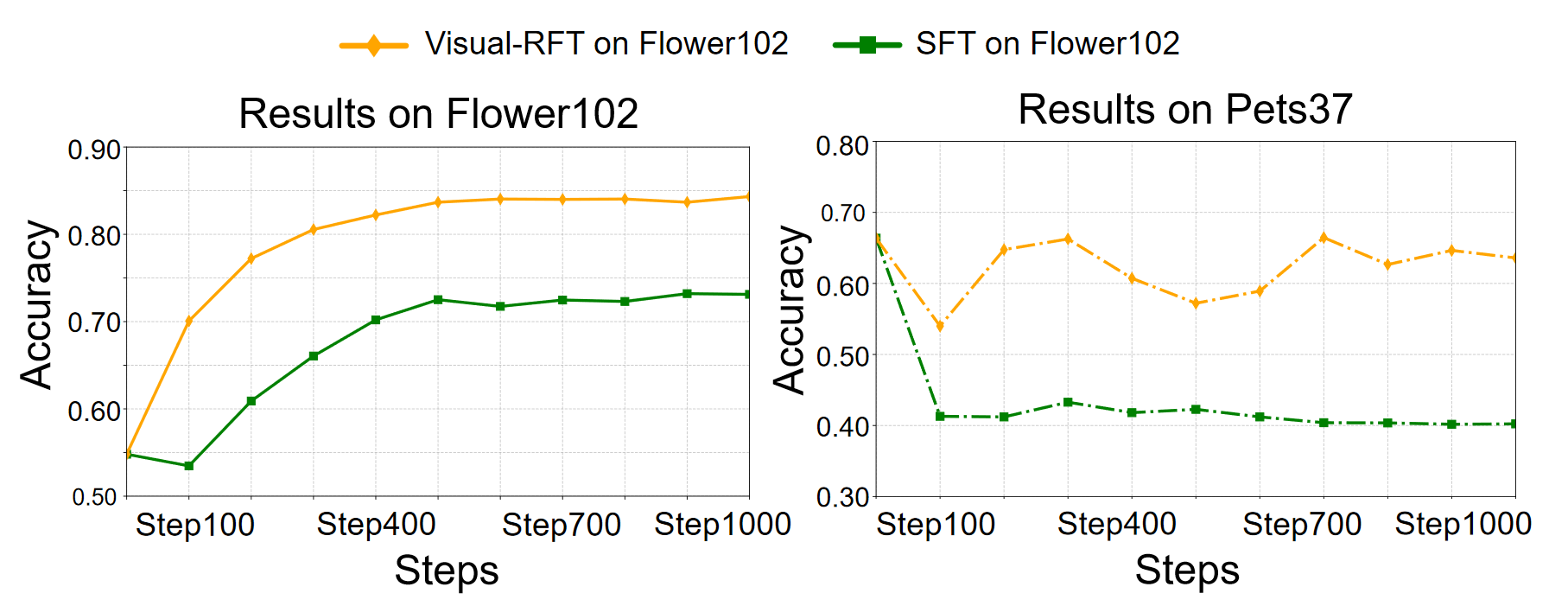}
    \end{center}
    \vspace{-2mm}
    \caption{\small \textbf{Ablation Study.} We evaluated \methodname and SFT across different steps and classification datasets. \methodname demonstrated superior performance and generalization capability compared to SFT.}
    \label{fig:abalation}
\end{figure}

\noindent \textbf{Generalization Ability.}
To further compare the performance and generalization capability of SFT and \methodname, we conduct the following ablation experiment. We use a 2-shot setting on the Flower102 dataset, training models separately with \methodname and SFT. The models are then tested on both the Flowers and Pets datasets. The results under different step settings are shown in Fig.~\ref{fig:abalation}.

In the left chart, the accuracy of both \methodname and SFT on the Flowers increases with the number of training steps, gradually reaching saturation. Notably, \methodname demonstrates a significant advantage over SFT when working with a small amount of data. In contrast, in the right chart, the accuracy of SFT on the Pets dataset drops sharply, falling well below the baseline. This decline is due to the model being trained on the Flower dataset, highlighting SFT’s limited generalization ability. Meanwhile, \methodname maintains its ability to distinguish fine-grained categories on the Pets dataset, showcasing its strong generalization performance even when the training and testing data domains differ.


\noindent \textbf{Reward Design.} To assess our reward design (~\ref{sec:methoda_reward}), we replace the F1 score-based reward with an EM-based alternative for both agentic search and coding.
As shown in Tab.~\ref{tab:ablation_study}, while EM-based training offers a slight improvement over the baseline, it consistently underperforms F1-based training.
This underscores the benefits of using the F1 score as a training signal: it provides smoother, more informative gradients by accounting for partial correctness, is more tolerant of linguistic variation, and facilitates stable training. Consequently, the F1 score-based reward is our default reward design.

\begin{table}[t]
\caption{\small \textbf{Ablation Study} on the reward design of multimodal agent.}
\label{tab:ablation_study}
\setlength{\tabcolsep}{4.7pt}
\renewcommand{\arraystretch}{1.0}
\scalebox{1.}{
\begin{tabular}{l|c|cc|cc|cc}
\toprule
\multirow{2}{*}{Task} &
\multirow{2}{*}{Reward} & 
\multicolumn{2}{c|}{\textbf{Simple}} &
\multicolumn{2}{c|}{\textbf{Hard}} &
\multicolumn{2}{c}{\textbf{Avg}} \\
\cmidrule(lr){3-4} \cmidrule(lr){5-6} \cmidrule(lr){7-8}
& & \textit{F1} & \textit{EM} & \textit{F1} & \textit{EM} & \textit{F1} & \textit{EM} \\
\midrule
MAT & EM & 53.50 & 45.71 & 43.00 & 23.85 & 46.68  & 31.50 \\
Coding & \cellcolor[HTML]{DAEFF9} F1 
  & \cellcolor[HTML]{DAEFF9} 60.10 & \cellcolor[HTML]{DAEFF9} 51.43 
  & \cellcolor[HTML]{DAEFF9} 45.60 & \cellcolor[HTML]{DAEFF9} 25.38 
  & \cellcolor[HTML]{DAEFF9} 50.68 & \cellcolor[HTML]{DAEFF9} 34.50 \\
\midrule

MAT & EM 
  & 59.23 & 53.33 
  & 53.37 & 41.33 
  & 56.30 & 47.33 \\
Search  & \cellcolor[HTML]{DAEFF9} F1 
  & \cellcolor[HTML]{DAEFF9} 71.78 & \cellcolor[HTML]{DAEFF9} 66.67 
  & \cellcolor[HTML]{DAEFF9} 55.77 & \cellcolor[HTML]{DAEFF9} 44.00 
  & \cellcolor[HTML]{DAEFF9} 63.77 & \cellcolor[HTML]{DAEFF9} 55.33 \\
\bottomrule
\end{tabular}
}
\vspace{-4pt}
\end{table}

\begin{table}[t]
\caption{\small \textbf{Ablation Study} with Direct inference/CoT/RAG baselines on \searchname.}
\vspace{-2pt}
\label{tab:ablation_study_cot_rag}
\setlength{\tabcolsep}{2.7pt}
\renewcommand{\arraystretch}{1.0}
\scalebox{1.}{
\begin{tabular}{l|c|cc|cc|cc}
\toprule
\multirow{2}{*}{Task} &
\multirow{2}{*}{Method} & 
\multicolumn{2}{c|}{\textbf{Simple}} &
\multicolumn{2}{c|}{\textbf{Hard}} &
\multicolumn{2}{c}{\textbf{Avg}} \\
\cmidrule(lr){3-4} \cmidrule(lr){5-6} \cmidrule(lr){7-8}
& & \textit{F1} & \textit{EM} & \textit{F1} & \textit{EM} & \textit{F1} & \textit{EM} \\
\midrule
Qwen2.5 & Direct & 57.54 & 50.67 & 33.11 & 26.67 & 45.32 & 38.67 \\
\quad -VL & CoT & 37.84 & 30.67 & 31.03 & 24.00 & 34.43 & 27.33 \\
\quad -3B & RAG & 49.42 & 45.33 & 39.07 & 32.00 & 44.25 & 38.67 \\
 & \methodname & \cellcolor[HTML]{DAEFF9} 56.41 & \cellcolor[HTML]{DAEFF9} 50.67 &
\cellcolor[HTML]{DAEFF9} 45.55 & \cellcolor[HTML]{DAEFF9} 36.00 & 
\cellcolor[HTML]{DAEFF9} 50.98 & \cellcolor[HTML]{DAEFF9} 43.33 \\
\midrule
Qwen2.5 & Direct & 67.40 & 61.33 & 39.59 & 32.00 & 53.49 & 46.67 \\
\quad -VL & CoT & 57.57 & 49.33 & 46.65 & 32.00 & 52.11 & 40.67 \\
\quad -7B & RAG & 59.14 & 56.00 & 42.44 & 36.00 & 50.79 & 46.00 \\
 & \methodname & \cellcolor[HTML]{DAEFF9} 71.78 & \cellcolor[HTML]{DAEFF9} 66.67 & 
\cellcolor[HTML]{DAEFF9} 55.77 & \cellcolor[HTML]{DAEFF9} 44.00 &
\cellcolor[HTML]{DAEFF9} 63.77 & \cellcolor[HTML]{DAEFF9} 55.33 \\
\bottomrule
\end{tabular}
}
\vspace{-4pt}
\end{table}

\noindent \textbf{Comparison with RAG and CoT baselines.}\label{sec:ablation_arft}
\methodname achieves a substantial performance improvement over the baseline on \searchname. To further evaluate model capabilities, we compare it with several traditional approaches, including Retrieval-Augmented Generation (RAG) and Chain-of-Thought (CoT). As shown in Tab.~\ref{tab:ablation_study_cot_rag}, CoT facilitates reasoning in multi-hop question answering by encouraging step-by-step inference.
However, it lacks access to external knowledge, limiting its ability to answer fact-based questions. In contrast, RAG can retrieve relevant external information but struggles with reasoning and problem decomposition, making it less effective on \searchname.
\methodname overcomes the limitations of both approaches by integrating strong reasoning, task decomposition, and tool-use abilities, significantly outperforming both RAG and CoT.

Additionally, we observe that methods based on CoT prompting or standard RAG not only fail to improve performance, but also in some cases, lead to degradation. This suggests that for base models without task-specific adaptation, directly injecting external information (e.g., content retrieved via RAG) can be ineffective or even counterproductive due to information interference.
In contrast, models trained with \methodname learn not only how to invoke tools, but also how to selectively utilize relevant external information to support complex reasoning.

\begin{table}[t]
    \centering
    \caption{Ablation study of different coding reward designs on MAT-Coding.}
    \label{tab:ablation_code_reward}
    \setlength{\tabcolsep}{14pt}
    \renewcommand{\arraystretch}{0.9}
    \begin{tabular}{lcc}
        \toprule
        \multirow{2}{*}{\textbf{Method}} 
            & \multicolumn{2}{c}{\textbf{MAT-Coding}} \\
        \cmidrule(lr){2-3}
            & \textbf{F1} & \textbf{EM} \\
        \midrule
        Baseline (Qwen2.5-VL-7B)  
            & 32.12 & 21.50 \\
        \midrule
        \quad + Visual-RFT++ (Ours) 
            & 50.68 & 34.50 \\
        \quad + w/o Code Reward       
            & 35.64 & 23.50 \\
        \quad + Execution-success Reward
            & 51.31 & 34.00 \\
        \quad + Strict Code Reward
            & 28.37 & 19.50 \\
        \bottomrule
    \end{tabular}
\vspace{-8pt}
\end{table}

\textbf{Ablation on the Coding Reward.}
We further compare different intermediate reward designs for MAT-Coding in Tab.~\ref{tab:ablation_code_reward}. Removing the code reward decreases F1 and EM by 15.04 and 11.00 points, respectively, demonstrating the importance of intermediate supervision for learning tool-based actions. The stricter execution-success reward achieves comparable performance to our code-block reward ($+0.63$ F1 and $-0.50$ EM). The small difference is consistent with the coding behavior diagnostics above (Sec.~\ref{sec:C4}), which show that most generated code actions are already executable. By contrast, encouraging similarity to a reference implementation (Strict Code Reward) substantially degrades performance, indicating that such constraints may penalize functionally valid alternative implementations.

\begin{table}[t]
    \centering
    \caption{Controlled comparison with tool-enabled baselines on MAT. Tool-enabled methods use the same external environment within each task.}
    \label{tab:tool_enabled_baselines}
    \footnotesize
    \setlength{\tabcolsep}{10pt}
    \renewcommand{\arraystretch}{1.08}
    \begin{tabular}{@{}llcc@{}}
        \toprule
        \textbf{Backbone} & \textbf{Method} & \textbf{F1} & \textbf{EM} \\
        \midrule
        \multicolumn{4}{c}{\textbf{MAT-Search}} \\
        \midrule
        \multirow{5}{*}{Qwen2.5-VL-7B}
        & Direct              & 53.49 & 46.67 \\
        & + ReAct             & 55.73 & 48.00 \\
        & + RAG               & 50.79 & 46.00 \\
        & + IRCoT             & 55.27 & 47.33 \\
        & + Visual-RFT++      & \textbf{63.77} & \textbf{55.33} \\
        \addlinespace[2pt]
        \multirow{2}{*}{GPT-4o}
        & Direct              & 61.08 & 52.00 \\
        & + Search tool       & \textbf{72.07} & \textbf{66.00} \\
        \midrule
        \multicolumn{4}{c}{\textbf{MAT-Coding}} \\
        \midrule
        \multirow{3}{*}{Qwen2.5-VL-7B}
        & Direct              & 32.12 & 21.50 \\
        & + Python execution  & 37.82 & 25.50 \\
        & + Visual-RFT++      & \textbf{50.68} & \textbf{34.50} \\
        \addlinespace[2pt]
        \multirow{2}{*}{GPT-4o}
        & Direct              & 34.41 & 23.50 \\
        & + Python execution  & \textbf{58.99} & \textbf{37.00} \\
        \bottomrule
    \end{tabular}
\vspace{-8pt}
\end{table}

\textbf{Controlled Comparison with Tool-Enabled Baselines.}
To distinguish the effect of reinforcement fine-tuning from that of tool access alone, we compare Visual-RFT++ with controlled tool-enabled baselines under the same external environments. On MAT-Search, Visual-RFT++ outperforms the strongest Qwen2.5-VL-7B tool-enabled baseline, ReAct, by 8.04 F1 and 7.33 EM points. On MAT-Coding, it exceeds Python/code-execution prompting by 12.86 F1 and 9.00 EM points. Equipping GPT-4o with the same tools also produces substantial improvements, reflecting its stronger inherent tool-using capabilities. The controlled same-backbone comparisons show that tool access alone does not account for the gains of Visual-RFT++; reinforcement fine-tuning further improves how the model invokes tools and uses their outputs.

\textbf{Transferability Across Backbones.}
To examine whether the improvements are specific to Qwen2.5-VL, we further apply Visual-RFT++ to the stronger Qwen3.5-9B backbone~\cite{qwen35blog}. As shown in Tab.~\ref{tab:different_backbone}, Visual-RFT++ improves its MAT-Search performance from 62.33/56.00 to 70.42/62.67 F1/EM, corresponding to gains of 8.09 F1 and 6.67 EM points. Together with the improvements of 10.28 F1 and 8.66 EM on Qwen2.5-VL-7B~\cite{bai2025qwen25vl}, these results support the transferability of Visual-RFT++ across the two evaluated backbones.

\begin{table}[t]
    \centering
    \caption{MAT-Search results with different backbones. The best result within each backbone is highlighted in bold.}
    \label{tab:different_backbone}
    \footnotesize
    \setlength{\tabcolsep}{10pt}
    \renewcommand{\arraystretch}{1.1}
    \begin{tabular}{llcc}
        \toprule
        \textbf{Backbone} & \textbf{Method} & \textbf{F1} & \textbf{EM} \\
        \midrule
        \multirow{2}{*}{Qwen2.5-VL-7B}
        & Baseline          & 53.49 & 46.67 \\
        & + Visual-RFT++    & \textbf{63.77} & \textbf{55.33} \\
        \midrule
        \multirow{2}{*}{Qwen3.5-9B}
        & Baseline          & 62.33 & 56.00 \\
        & + Visual-RFT++    & \textbf{70.42} & \textbf{62.67} \\
        \bottomrule
    \end{tabular}
\end{table}

\textbf{Baseline Comparison Fairness.}
To control for the backbone difference in Tab.~\ref{tab:multihopqa}, we retrain Search-R1~\cite{jin2025search-r1} on the same Qwen2.5-VL-7B-Instruct backbone~\cite{bai2025qwen25vl} using its original text-based training data. As shown in Tab.~\ref{tab:same_backbone_search_r1}, Visual-RFT++ achieves 48.49 F1 and 37.84 EM, outperforming the same-backbone Search-R1 by 6.13 F1 and 7.27 EM points. Notably, before reinforcement fine-tuning, Qwen2.5-VL-7B has a lower average EM than the text-only Qwen2.5-7B backbone used in Tab.~VIII (11.93 vs.\ 14.60). These results reduce the potential confounding effect of backbone differences in the original comparison.

\begin{table}[t]
    \centering
    \caption{Same-backbone comparison averaged over four text-only multi-hop QA benchmarks.}
    \label{tab:same_backbone_search_r1}
    \footnotesize
    \setlength{\tabcolsep}{10pt}
    \renewcommand{\arraystretch}{1.1}
    \begin{tabular}{lcc}
        \toprule
        \textbf{Method} & \textbf{F1} & \textbf{EM} \\
        \midrule
        Baseline (Qwen2.5-VL-7B) & 19.19 & 11.93 \\
        \quad + Search-R1         & 42.36 & 30.57 \\
        \quad + Visual-RFT++      & \textbf{48.49} & \textbf{37.84} \\
        \bottomrule
    \end{tabular}
\end{table}

\textbf{Sensitivity to GRPO Hyperparameters.}
We first examine the GRPO group size $G$ and KL coefficient $\beta$ on both agentic tasks. As shown in Tab.~\ref{tab:agentic_hyperparameters}, increasing $G$ from 4 to 8 improves F1 by 2.23 points on MAT-Search and 2.73 points on MAT-Coding, whereas increasing it further to 16 provides relatively modest additional gains. We therefore use $G=8$ to balance performance and rollout cost. The results are also stable across $\beta\in\{0.02,0.04,0.08\}$, with F1 varying by at most 0.65 points on MAT-Search and 0.39 points on MAT-Coding. We use $\beta=0.04$ by default.

\begin{table}[t]
    \centering
    \caption{Sensitivity analysis of the GRPO group size $G$ and KL coefficient $\beta$ on MAT-Search and MAT-Coding.}
    \label{tab:agentic_hyperparameters}
    \setlength{\tabcolsep}{6pt}
    \renewcommand{\arraystretch}{1.15}
    \begin{tabular}{lcccc}
        \toprule
        \multirow{2}{*}{\textbf{Method}}
            & \multicolumn{2}{c}{\textbf{MAT-Search}}
            & \multicolumn{2}{c}{\textbf{MAT-Coding}} \\
        \cmidrule(lr){2-3}
        \cmidrule(lr){4-5}
            & \textbf{F1} & \textbf{EM}
            & \textbf{F1} & \textbf{EM} \\
        \midrule
        Baseline (Qwen2.5-VL-7B)
            & 53.49 & 46.67 & 32.12 & 21.50 \\
        \midrule
        Visual-RFT++ ($G=8,\ \beta=0.04$)
            & 63.77 & 55.33 & 50.68 & 34.50 \\
        Visual-RFT++ ($G=4,\ \beta=0.04$)
            & 61.54 & 53.33 & 47.95 & 32.50 \\
        Visual-RFT++ ($G=16,\ \beta=0.04$)
            & 64.03 & 56.67 & 51.74 & 35.00 \\
        Visual-RFT++ ($G=8,\ \beta=0.02$)
            & 63.58 & 55.33 & 50.67 & 34.00 \\
        Visual-RFT++ ($G=8,\ \beta=0.08$)
            & 63.12 & 54.67 & 51.06 & 35.50 \\
        \bottomrule
    \end{tabular}
\end{table}

\textbf{Sensitivity to the IoU Threshold.}
We further vary the IoU threshold $\tau$ under the 8-shot few-shot detection setting of Tab.~\ref{tab:few-shot_coco}. As shown in Tab.~\ref{tab:iou_threshold}, $\tau=0.5$ achieves the highest mAP. A lower threshold admits more low-quality box matches, whereas a higher threshold produces fewer positive matches and sparser supervision. We therefore use $\tau=0.5$ in the main experiments.

\begin{table}[t]
    \centering
    \caption{Sensitivity analysis of the IoU threshold $\tau$ under the few-shot detection setting of Tab.~\ref{tab:few-shot_coco}.}
    \label{tab:iou_threshold}
    \setlength{\tabcolsep}{6pt}
    \resizebox{0.47\textwidth}{!}{
    \begin{tabular}{lccccccccc}
        \toprule
        \textbf{Method} & \textbf{mAP} & \rotatebox{90}{\textbf{bus}} & \rotatebox{90}{\textbf{train}}
        & \rotatebox{90}{\textbf{fire hydrant}} & \rotatebox{90}{\textbf{stop sign}}
        & \rotatebox{90}{\textbf{cat}} & \rotatebox{90}{\textbf{dog}} & \rotatebox{90}{\textbf{bed}} & \rotatebox{90}{\textbf{toilet}} \\
        \midrule
        SFT baseline
            & 30.2 & 25.8 & 35.1 & 29.4 & 21.9 & 44.5 & 39.0 & 22.6 & 23.5 \\
        Ours ($\tau=0.5$)
            & \textbf{47.4} & \textbf{36.2} & \textbf{47.9}
            & \textbf{50.4} & \textbf{47.7} & \textbf{65.2}
            & \textbf{57.0} & \textbf{30.4} & \textbf{44.0} \\
        Ours ($\tau=0.3$)
            & 45.1 & 33.6 & 45.2 & 48.7 & 44.3 & 62.8 & 54.9 & 28.4 & 42.4 \\
        Ours ($\tau=0.7$)
            & 43.6 & 31.2 & 43.5 & 47.6 & 42.1 & 62.8 & 53.4 & 27.1 & 40.6 \\
        \bottomrule
    \end{tabular}}
\end{table}

\textbf{Sensitivity to Reward Weights.}
We also vary the relative weights of the format, IoU, and confidence rewards by halving each component individually from the default $1{:}1{:}1$ setting. The default setting achieves 47.40 mAP, while reducing the format-, IoU-, or confidence-reward weight yields 45.73, 44.04, and 44.96 mAP, respectively. These results show that all three components contribute to performance, with the IoU and confidence rewards having a relatively larger influence on localization accuracy.

\textbf{Multi-Seed Evaluation.}
To evaluate the stability of the reward-design ablation, we repeat the EM- and F1-based reward experiments using three random seeds. As shown in Tab.~\ref{tab:multiseed_reward}, the multi-seed mean results remain close to the single-seed results in Tab.~\ref{tab:ablation_study}, with standard deviations ranging from 0.7 to 1.2.

\begin{table}[t]
    \centering
    \caption{Multi-seed evaluation of the EM- and F1-based reward settings. Results are reported as mean $\pm$ standard deviation over three random seeds.}
    \label{tab:multiseed_reward}
    \setlength{\tabcolsep}{9pt}
    \renewcommand{\arraystretch}{1.15}
    \begin{tabular}{llcc}
        \toprule
        \multirow{2}{*}{\textbf{Task}}
            & \multirow{2}{*}{\textbf{Reward}}
            & \multicolumn{2}{c}{\textbf{Evaluation Metric}} \\
        \cmidrule(lr){3-4}
            & & \textbf{F1} & \textbf{EM} \\
        \midrule
        \multirow{2}{*}{MAT-Coding}
            & EM & $46.34\pm0.8$ & $32.00\pm1.0$ \\
            & F1 & $\mathbf{50.86\pm0.9}$ & $\mathbf{35.00\pm1.0}$ \\
        \midrule
        \multirow{2}{*}{MAT-Search}
            & EM & $56.72\pm0.8$ & $48.00\pm0.7$ \\
            & F1 & $\mathbf{63.94\pm0.7}$ & $\mathbf{55.33\pm1.2}$ \\
        \bottomrule
    \end{tabular}
\end{table}
\section{Conclusion}

We introduce Visual Reinforcement Fine-Tuning++ (\methodname), a unified RFT framework that extends the GRPO-based reinforcement learning strategy to both visual perception and multimodal agentic reasoning. Through the combination of GRPO optimization and modular rule-based rewards, \methodname enables LVLMs to acquire visual understanding, grounded reasoning, and tool-use capabilities with minimal supervision.
Experimentally, \methodname delivers strong gains across fine-grained classification, few-shot detection, reasoning grounding, and open-vocabulary detection, consistently outperforming supervised fine-tuning.
It also equips models with effective multimodal agentic behaviors, leading to advanced performance on our Multimodal Agentic Tool Bench (\benchname) and strong generalization to multi-hop QA benchmarks.

\bibliographystyle{IEEEtran}
\bibliography{main}

\newpage
\begin{IEEEbiography}[{\includegraphics[width=1in,height=1.25in,clip,keepaspectratio]{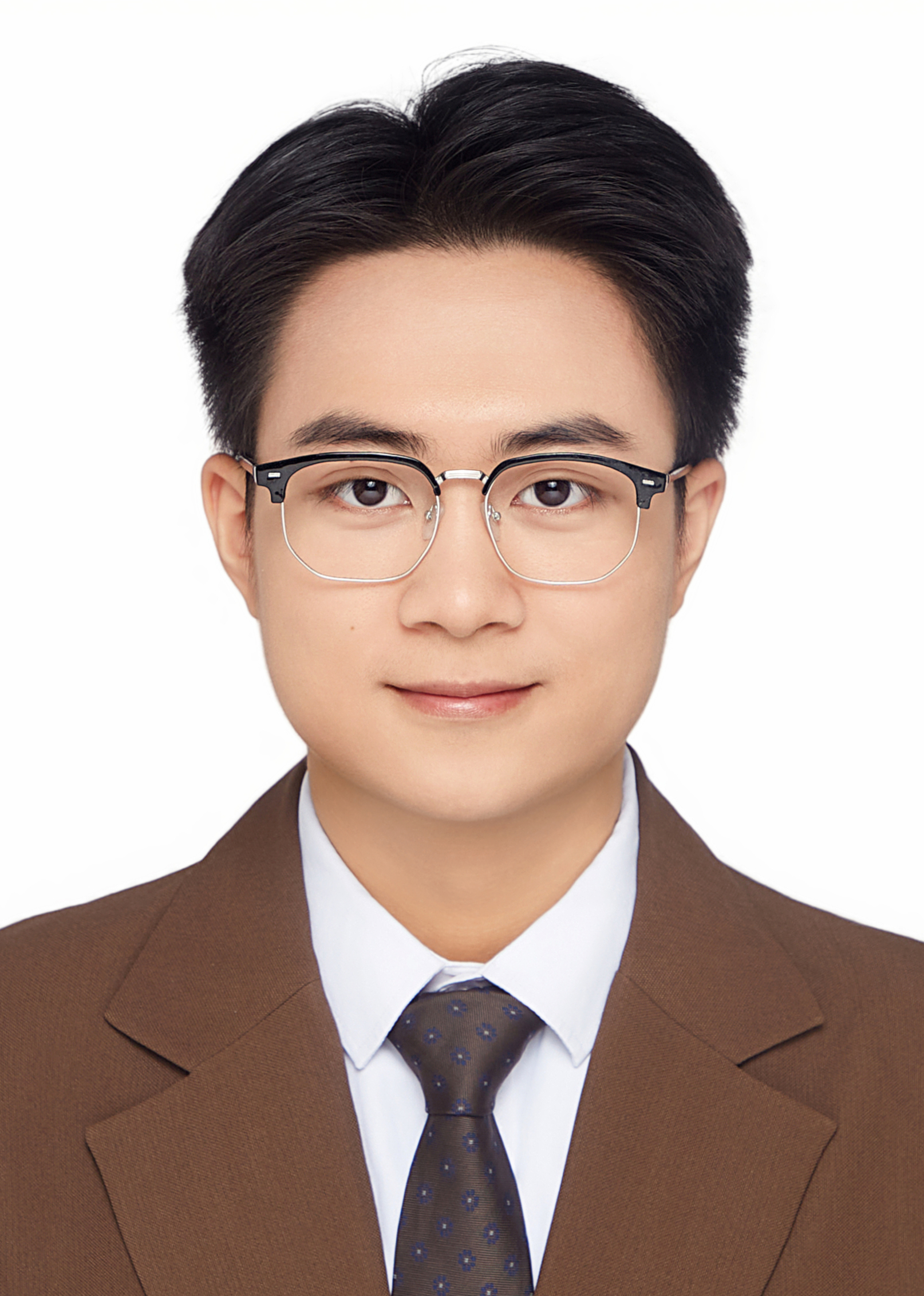}}] {Ziyu Liu} is a PhD student at Shanghai Jiao Tong University, under the supervision of Prof. Dahua Lin. He is currently a research intern at Shanghai Artificial Intelligence Laboratory, mentored by Yuhang Zang. His research interests focus on Large Vision Language Models, Reinforcement Fine-tuning, and Multimodal Agentic RL.
\end{IEEEbiography}

\begin{IEEEbiography}[{\includegraphics[width=1in,height=1.25in,clip,keepaspectratio]{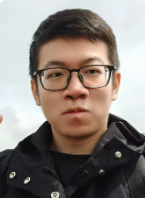}}] {Zeyi Sun} is a third-year PhD student in Information Engineering at Shanghai Jiao Tong University, under the supervision of Prof. Dahua Lin. He obtained his Bachelor’s degree in Computer Science from Beihang University in 2023. His research interests focus on Large Vision Language Models, Reinforcement Fine-tuning, and Multimodal Agentic RL.
\end{IEEEbiography}

\begin{IEEEbiography}
[{\includegraphics[width=1in,height=1.25in,clip,keepaspectratio]{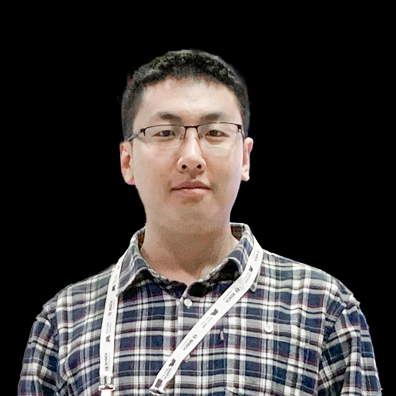}}] {Xiaoyi Dong} is a Postdoctoral Fellow at MMLab, The Chinese University of Hong Kong. He also serves as a part-time consultant researcher at Shanghai AI Laboratory. He obtained his Ph.D. degree at the University of Science and Technology of China in 2023. He has published more than 40 papers in top conferences and journals, such as CVPR/ICCV/ECCV, and has received than 8000 citations according to Google Scholar. His research interests focus on multimodel models and general representation learning. He is the recipient of the CAS President's Scholarship.
\end{IEEEbiography}

\begin{IEEEbiography}[{\includegraphics[width=1in,height=1.25in,clip,keepaspectratio]{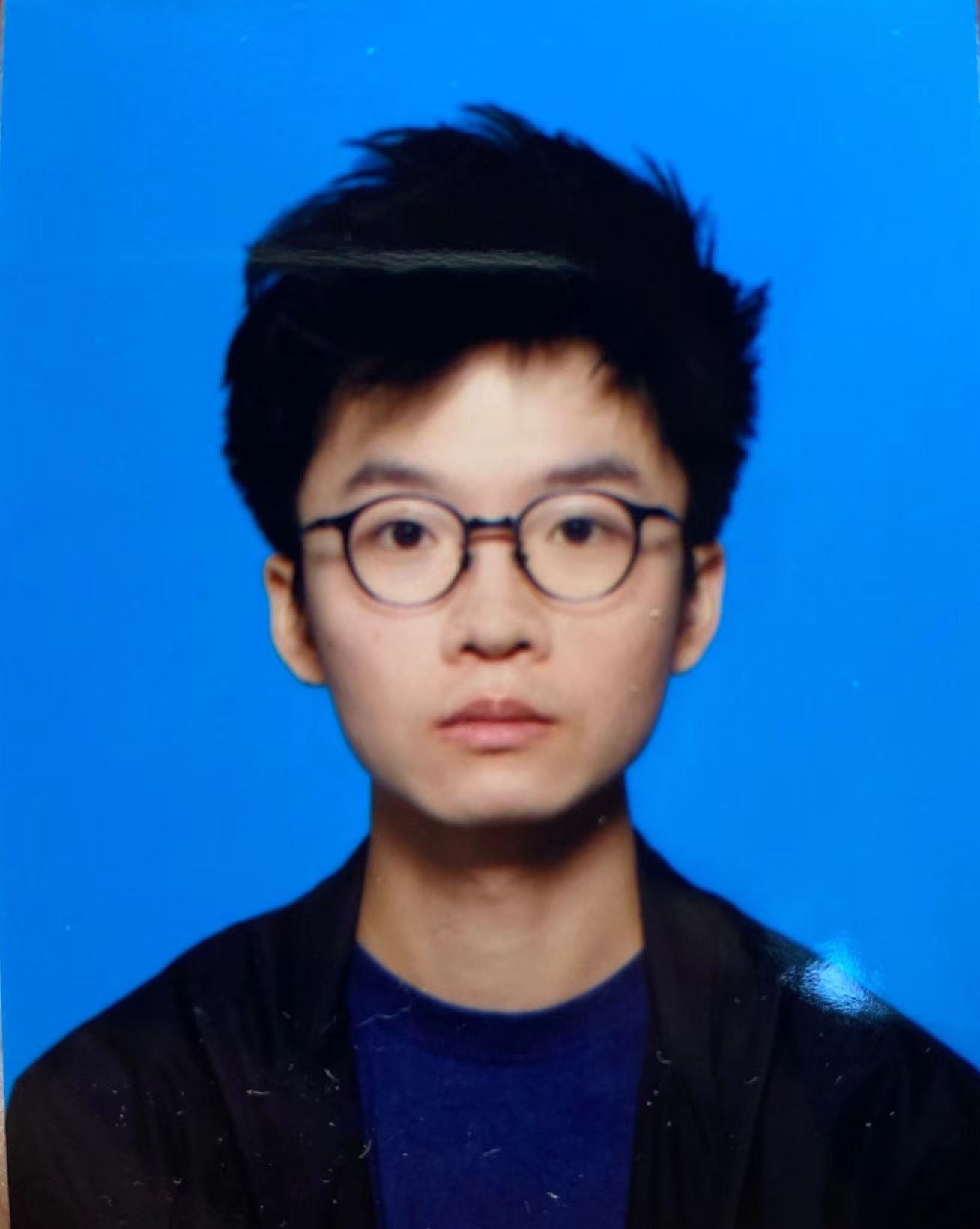}}] {Yuhang Cao} is a researcher at the Shanghai Artificial Intelligence Laboratory. He received his Ph.D. from The Chinese University of Hong Kong in 2024, under the supervision of Prof. Dahua Lin. He has authored over 30 papers in top-tier conferences and journals, including CVPR, ICCV, and ECCV, and his work has received more than 6,000 citations according to Google Scholar. His research interests lie primarily in vision language models and image object detection.
\end{IEEEbiography}

\begin{IEEEbiography}[{\includegraphics[width=1in,height=1.25in,clip,keepaspectratio]{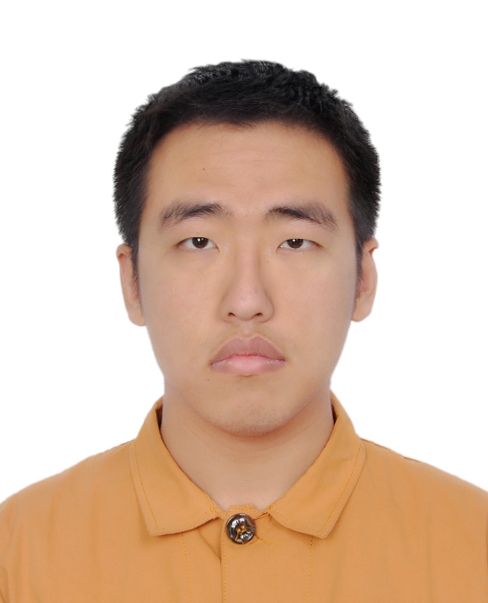}}] {Haodong Duan} is a Post-Doctoral Researcher at Shanghai AI Laboratory. 
He received his Bachelor's degree from Peking University and his Ph.D. from The Chinese University of Hong Kong. 
His research experience spans video understanding, large language models, and multimodal learning. 
He has published 30+ papers in top-tier conferences in computer vision and artificial intelligence, 
including CVPR, ECCV, ICCV, ACL, and NeurIPS, with more than 6,000 citations.
Recently, he mainly works on the evaluation of large multimodal models.
\end{IEEEbiography}

\begin{IEEEbiography}[{\includegraphics[width=1in,height=1.25in,clip,keepaspectratio]{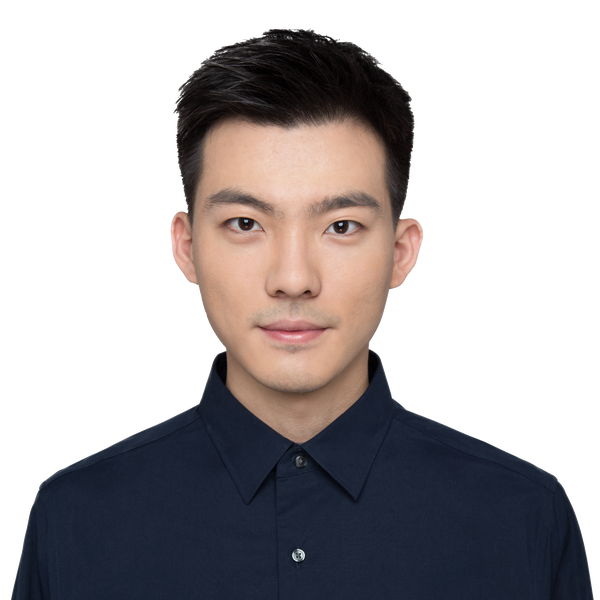}}] {Jiaqi Wang} is currently a Research Scientist and Team Leader at Shanghai AI Laboratory, and an Adjunct Ph.D. Supervisor at Shanghai Jiao Tong University. He received his Ph.D. in 2021 from the Multimedia Laboratory at The Chinese University of Hong Kong (MMLab@CUHK), under the supervision of Prof. Dahua Lin and in close collaboration with Prof. Chen Change Loy. His research interests span visual perception, vision-language models, and the creation of benchmarks and datasets. He has published over 50 papers in top-tier venues, including TPAMI, CVPR, ICCV, ECCV, NeurIPS, ICML, and ICLR, with more than 15,000 citations. His work has received broad recognition: two papers were named among the Most Influential Papers at ECCV 2024; one was selected as a Most Influential Paper at NeurIPS 2024; and another was nominated as an Award Candidate at CVPR 2023. He was also listed among the Stanford/Elsevier Top 2\% Scientists in 2024. Jiaqi is the lead researcher of the InternLM-XComposer series of large multimodal models, which have collectively been downloaded over 4 million times and at one point ranked 5th out of more than 750,000 models on HuggingFace Trending.
\end{IEEEbiography}

\begin{IEEEbiography}[{\includegraphics[width=1in,height=1.25in,clip,keepaspectratio]{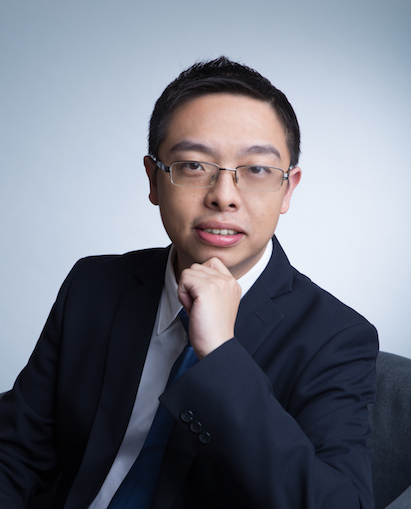}}]{Dahua Lin} is an Associate Professor at the Department of Information Engineering, the Chinese University of Hong Kong, and the Director of CUHK-SenseTime Joint Laboratory. He received the B.Eng. degree from the University of Science and Technology of China (USTC) in 2004, the M. Phil. degree from the Chinese University of Hong Kong (CUHK) in 2006, and the Ph.D. degree from Massachusetts Institute of Technology (MIT) in 2012. Prior to joining CUHK, he served as a Research Assistant Professor at Toyota Technological Institute at Chicago, from 2012 to 2014. His research interests cover computer vision and machine learning.
\end{IEEEbiography}

\begin{IEEEbiography}[{\includegraphics[width=1in,height=1.25in,clip,keepaspectratio]{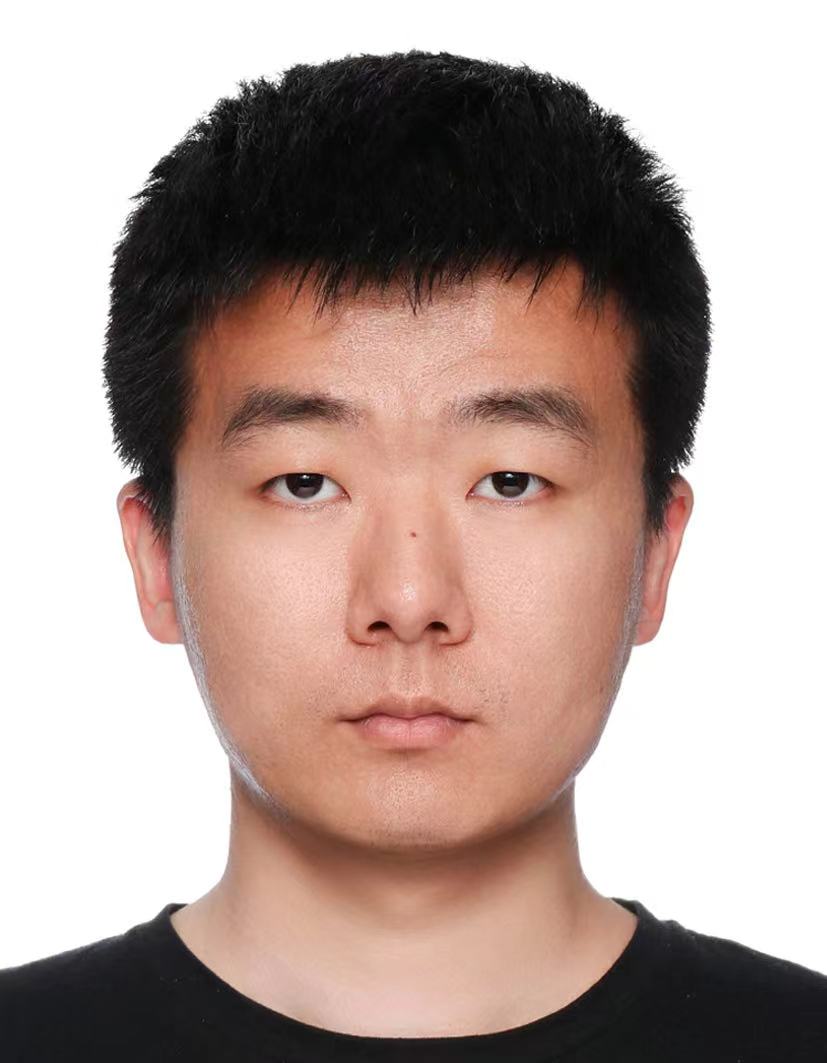}}]{Yuhang Zang} is a researcher at Shanghai Artificial Intelligence Laboratory. He received the PhD degree from Nanyang Technological University in 2023, under the supervision of Prof. Chen Change Loy. Before that, he received the B.Eng degree from the University of Electronic Science and Technology of China in 2019. His research interests cover vision-language models and large language models.
\end{IEEEbiography}

\end{document}